\documentclass{article}

\usepackage{arxiv}

\usepackage[utf8]{inputenc} 
\usepackage[T1]{fontenc}    
\usepackage{hyperref}       
\usepackage{url}            
\usepackage{booktabs}       
\usepackage{amsfonts,amsmath}       
\usepackage{nicefrac}       
\usepackage{microtype}      
\usepackage{lipsum}
\usepackage{graphicx}
\usepackage{bm}
\graphicspath{ {./Figures/} }
\usepackage{amssymb}
\usepackage{array}
\usepackage{mathtools}
\usepackage{color}
\usepackage{tabularx,booktabs}
\newcommand{\smallcdot}[0]{\! \cdot \!}
\usepackage{algorithm,algorithmicx,algpseudocode}
\usepackage{enumitem}
\definecolor{mblue}{RGB}{57,106,177}
\hypersetup{
    citecolor=blue,
	colorlinks=true,
	linkcolor=blue,
	filecolor=blue,      
	urlcolor=blue
}
\title{Stochastic stiffness identification and response estimation of Timoshenko beams via physics-informed Gaussian processes}

\author{
  Gledson Rodrigo Tondo$^{\star}$ \\
  Bauhaus-Universität Weimar\\
  Weimar, Germany \\
  \And
  Sebastian Rau \\
  Bauhaus-Universität Weimar\\
  Weimar, Germany \\
  \And
  Igor Kavrakov \\
  University of Cambridge\\
  Cambridge, United Kingdom \\
  \And
  Guido Morgenthal \\
  Bauhaus-Universität Weimar\\
  Weimar, Germany \\
}

\begin{document}
\maketitle
\begin{abstract}
Machine learning models trained with structural health monitoring data have become a powerful tool for system identification. This paper presents a physics-informed Gaussian process (GP) model for Timoshenko beam elements. The model is constructed as a multi-output GP with covariance and cross-covariance kernels analytically derived based on the differential equations for deflections, rotations, strains, bending moments, shear forces and applied loads. Stiffness identification is performed in a Bayesian format by maximising a posterior model through a Markov chain Monte Carlo method, yielding a stochastic model for the structural parameters. The optimised GP model is further employed for probabilistic predictions of unobserved responses. Additionally, an entropy-based method for physics-informed sensor placement optimisation is presented, exploiting heterogeneous sensor position information and structural boundary conditions built into the GP model. Results demonstrate that the proposed approach is effective at identifying structural parameters and is capable of fusing data from heterogeneous and multi-fidelity sensors. Probabilistic predictions of structural responses and internal forces are in closer agreement with measured data. We validate our model with an experimental setup and discuss the quality and uncertainty of the obtained results. The proposed approach has potential applications in the field of structural health monitoring (SHM) for both mechanical and structural systems.
\end{abstract}

\section{Introduction}	\par
	The classical Timoshenko beam theory is one of the most applied models in modern engineering, especially for structures with a high depth-to-length ratio, or composite and sandwich beams. Recent advances in material technology and controlled manufacturing have increased the use of such structures. From a sustainability point of view, it is important that these systems are maintained and that their life-cycle is extended. Thus, a large number of sensors and computer technologies are commonly applied in structural health monitoring (SHM) systems.\par
\begin{figure}[h]
		\centering
		\includegraphics[width=0.48\textwidth]{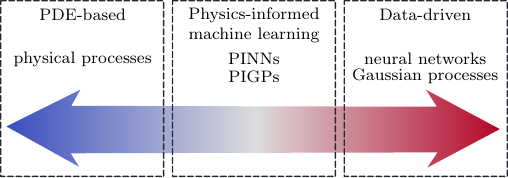}
		\caption{The three main types of modelling approaches. PDE-based models represent physical phenomena in terms of differential equations, while data-driven models can only reflect the properties contained in the training data. Physics-informed machine learning models are hybrid in nature, and fit training data according to their mathematical description provided by PDEs.}\label{fig_modelarrow}
	\end{figure}

	
	Traditionally, SHM depends on the understanding of the system's physical behaviour, generally described by partial differential equations (PDEs), and structural diagnostics relies on optimisation and model updating strategies~\cite{jahangiriProcedureEstimateMinimum2023,tacirogluEfficientModelUpdating2017, noever-castelosModelUpdatingWind2022,shanParametricIdentificationTimoshenkobeam2023,zapicoFiniteElementModel2003}. Advances in machine learning allowed for the use of data-driven methods that lack a physical description of a process or structure, such as neural networks (NNs) and Gaussian process (GP) regression. These have been extensively and successfully used as surrogate models (cf. eg.~\cite{parkEstimationInputParameters2002,fangDamageIdentificationResponse2011,gulianGaussianProcessRegression2022,haghighatPhysicsinformedDeepLearning2021}) to accelerate model updating, effectively replacing expensive numerical simulations. Despite their good predictive quality, the lack of a physical description can eventually lead to a poor performance, as the model is fully based on the data that it is trained with. Recently, physics-informed machine learning models have been introduced as a hybrid formulation to bridge the gap between PDE-based and data-driven strategies~\cite{karniadakisPhysicsinformedMachineLearning2021a}. These models can learn from data, but they do so while conforming to specific descriptions of physical phenomena that are built into the machine-learning framework using partial differential equations. A schematic of the three different approaches is shown in Fig.~\ref{fig_modelarrow}. Combinations of PDE-based models and neural networks have been employed to solve problems in material science~\cite{caiPhysicsInformedNeuralNetworks2021,henkesPhysicsInformedNeural2022,zhangAnalysesInternalStructures2022}, fluid dynamics~\cite{raissiPhysicsinformedNeuralNetworks2019,maoPhysicsinformedNeuralNetworks2020,jinNSFnetsNavierStokesFlow2021} and structural mechanics~\cite{tripathiPhysicsintegratedDeepLearning2023,abueiddaMeshlessPhysicsinformedDeep2021, yanFrameworkBasedPhysicsinformed2022,sharmaPhysicsinformedNeuralNetworks2021}. \par
    Gaussian process (GP) regression~\cite{rasmussenGaussianProcessesMachine2006} have gained popularity as a non-parametric probabilistic machine learning tool with a powerful learning framework that avoids overfitting by construction~\cite{maProbabilisticReconstructionSpatiotemporal2022,parussiniMultifidelityGaussianProcess2017, gregorySynthesisDataInstrumented2019, xiaoExtendedCoKrigingInterpolation2018}. They can be viewed as a single-layer neural network with an infinite number of neurons~\cite{nealPriorsInfiniteNetworks1996}. A physically-informed GP model is defined by a set of Gaussian processes with mean functions and covariance matrices derived based on a particular differential equation. The relation between different quantities defined by the PDE of interest is modelled by cross-covariance kernels~\cite{raissiMachineLearningLinear2017,alvarezLinearLatentForce2013a,sarkkaLinearOperatorsStochastic2011}. Physics informed GPs have been a popular tool to estimate the inverse solution of PDEs~\cite{raissiHiddenPhysicsModels2018,raissiMachineLearningLinear2017}, where the unknown model parameters are identified from noisy measurement data by incorporating them as a variable in the covariance kernels. This strategy has been applied to the system identification of Bernoulli beams~\cite{tondoPhysicsinformedGaussianProcess2022,gregorySynthesisDataInstrumented2019} and the Navier-Stokes equation parameters~\cite{raissiHiddenPhysicsModels2018}. In a forward approach, PIGPs have been employed for the dynamic force identification in structural systems~\cite{rogersLatentRestoringForce2022,nayekGaussianProcessLatent2019,alvarezLinearLatentForce2013a}, applied load reconstruction in slender structures and for aerodynamic analysis of long-span bridges~\cite{kavrakovDatadrivenAerodynamicAnalysis2022}. The probabilistic nature of Gaussian processes have also been used as a tool for sensor placement optimisation and online learning~\cite{durWeakConstraintGaussian2020,xuMultiobjectiveOptimizationSensor2022,tajnafoiVariationalGaussianProcess2021,parkGaussianProcessOnline2020}, determining optimal data collection locations via a greedy entropy minimization approach~\cite{krauseNearOptimalSensorPlacements2008}. To the best of the author's knowledge, this strategy has not yet been employed within a physics-informed framework.\par
    In this study, we introduce a physics-informed GP model based on the Timoshenko theory of static response of beams~\cite{timoshenkoLXVICorrectionShear1921}. The model is constructed as a multi-output GP with covariances and cross-covariances analytically derived based on the differential equation models for the physical quantities (deflections, rotations, strains, bending moments, shear forces and applied loads). This approach allows the model to be trained based of combinations of heterogeneous datasets (e.g. a combination of displacements and strains) while accounting for their correlation, defined by the cross-covariance kernels. In addition, the model also allows for training based on several datasets of the same response with different quality levels, by determining individual optimal noise values for each dataset. Bending and shear stiffness are assumed constant throughout the length of the structure and their identification is carried out in a Bayesian manner using a Markov chain Monte Carlo approach to sample from the posterior parameter distribution. This procedure results in stochastic models for the structural parameters, providing uncertainty bounds instead of point estimates as it is usually the case with the majority of the methods in literature. Predictions of unobserved responses are done in a fully Bayesian framework, taking into account model hyperparameter uncertainties. Additionally, a novel heterogeneous physics-informed sensor placement optimisation framework is presented. The method is based on information theory, and can capture cross-domain influence (e.g. the information gained about deflections at a certain location after placing a rotation sensor nearby). This also extends similar sensor sensor placement methodologies as it allows for the incorporation of boundary conditions prior to the optimisation. \par
    The paper is organized as follows: section~\ref{sec:TimoshenkoTheory} briefly reviews the Timoshenko beam model, with a particular view on how the bending and shear contributions to the response are linearly combined. Section~\ref{sec:GPModel} presents the derivations of the physics-informed GP model, along with the framework for parameter learning, stiffness identification and prediction strategy. Section~\ref{sec:SensorPlacement} defines the entropy minimization approach for sensor placement optimisation and discusses the advantages obtained from the physics-informed setting. To evaluate the model, section~\ref{sec:NumericalStudy} presents numerical studies comparing the novel and standard entropy-based sensor placement strategies and uses the results for stiffness identification and further predictions of model responses. Influences of noise and structural rigidity are also discussed. Validation of the novel method is given in section~\ref{sec:ExperimentalResults} with an experimental setup, where heterogeneous multi-fidelity data sets are used for the inverse and forward problems within the GP framework. Lastly, section~\ref{sec:conclusion} presents the conclusions of this study, where limitations and possible improvements are discussed. The implementations of the model can be found in the Github repository available at \href{https://github.com/gledsonrt/PIGPTimoshenkoBeam}{https://github.com/gledsonrt/PIGPTimoshenkoBeam}.
	\section{Timoshenko beam theory} \label{sec:TimoshenkoTheory} \par
    \begin{figure*}[h]
		\centering
		\includegraphics[width=1\textwidth]{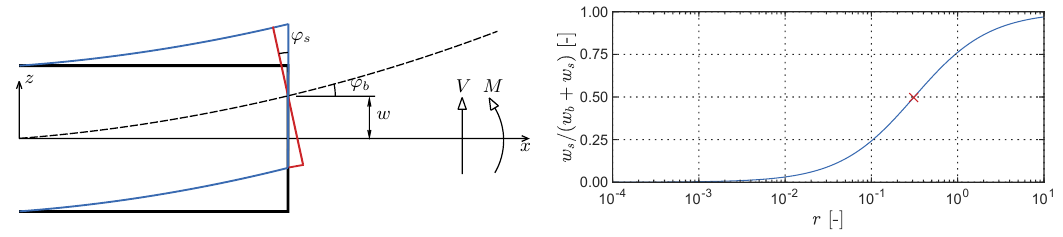}
		\caption{Left: The Timoshenko beam model (blue) accounts for shear deformations of the cross-section. When the rotation due to shear $\varphi_s = -V / kGA$ is negligible the traditional Euler-Bernoulli beam model (red) is recovered. Right: Relative deflection contribution due to shear effects, as a function of the structural rigidity. The inflexion point at $r=0.3125$, for a simply supported beam under UDL, corresponds to the case where 50\% of the total deflection $w$ is due to shear effects.}\label{fig:FigBeam}		
	\end{figure*}		
    Traditional static beam models define the response of a line-like structure as a function of an externally applied load. The Timoshenko (TB) beam theory~\cite{timoshenkoLXVICorrectionShear1921} considers the total deflection response $w(x)$ along the length-wise position $x$ as a contribution from bending $w_b$ and shear $w_s$ effects (see Fig.~\ref{fig:FigBeam}, left), such that	
	\begin{equation}
		w(x) = w_{b} + w_{s}.
		\label{eq:TimoTotalDeflection}
	\end{equation}	
	\noindent The deflection due to bending, assuming constant stiffness along the length of the beam, can be obtained from the applied load $q(x)$ as	
	\begin{equation}
		q(x) =  EI \frac{d^4 w_b}{dx^4} , 
		\label{eq:Load}
	\end{equation}	
	\noindent where $E$ is the modulus of elasticity and $I$ is the second moment of area of the cross-section. Assuming small deflections, the angle of rotation due to bending effects $\varphi_b$ normal to the mid-surface of the beam is obtained by	
	\begin{equation}
		\varphi_b(x) = \frac{d w_b}{dx}.
		\label{eq:TimoRotation}
	\end{equation}	\par
    \noindent Bending moments $M$ in the linear elastic range are related to the deflections via the bending stiffness $EI$ by:	
	\begin{equation}
		M(x) = EI \frac{d^2 w_b}{dx^2}.
		\label{eq:TimoMoment}
	\end{equation}	
	\noindent The shear forces are obtained through the derivative of the bending moment w.r.t. the spatial coordinate, 	
	\begin{equation}
		V(x) = \frac{d M}{dx} =  EI \frac{d^3 w_b}{dx^3}.
		\label{eq:TimoShear}
	\end{equation}	
	The total rotation of the cross-section, including the contribution from shear effects $\varphi_s$~\cite{timoshenkoLXVICorrectionShear1921}, is calculated as:	
	\begin{equation}
		\varphi(x) = \varphi_b + \varphi_s = \frac{d w_b}{dx} - \frac{V}{kGA},
		\label{eq:TimoFullRot}
	\end{equation}	
	\noindent where $G$ is the shear modulus, $A$ is the cross-section area, and $k$ is the Timoshenko shear coefficient, which accounts for the differences between the average and the exact shear supported by a cross-section of arbitrary geometry. The deflection $w$ is obtained integrating $\varphi$ as	
	\begin{equation}
		w(x) = \int \varphi dx.
		\label{eq:TimoDef}
	\end{equation}	
	\noindent The strain $\epsilon$ is linear across the vertical direction $z$ in the cross-section and can be obtained from the rotations as	
	\begin{equation}
		\epsilon(x,z) = - z \frac{d \varphi}{dx}.
		\label{eq:TimoStrain}
	\end{equation}
	Boundary conditions (BCs) are accounted for by enforcing the responses at specific positions. For instance, at simple supports located at $x = x_{\mathrm{BC}}$ that allows for structural rotation, the respective boundary condition is $w(x_{\mathrm{BC}}) = 0$ m, or in case of supports that restrict displacements and rotations, $w(x_{\mathrm{BC}}) = 0$ m and $\varphi(x_{\mathrm{BC}}) = 0$ rad.	
	Following Timoshenko's beam relations, the total response of the model is a function of both the bending stiffness $EI$ and the shear stiffness $kGA$. This dependency can be described by a rigidity factor $r$, calculated as~\cite{prathapLockingFiniteelementAnalysis1991, prathapScienceFEAPatterns1999}:	
	\begin{equation}
		r = \frac{3 EI}{L^2kGA},
		\label{eq:rigidityFact}
	\end{equation}	
	\noindent where $L$ is the beam length. If the shear stiffness $kGA$ tends to infinity and  $r \ll 1$, the deflection contribution due to shear is negligible and the traditional Euler-Bernoulli beam model is obtained. For example, in the case of a simply supported beam under uniformly distributed load (UDL), at $r=0.3125$ the combination of material and geometrical properties results in equal deflection contributions from shear and bending effects (c.f. Fig.~\ref{fig:FigBeam}, right). 	\par
	\section{Physics-informed GP model} \label{sec:GPModel} \par
	\subsection{Problem statement}\par
	Consider a heterogeneous data set $\mathcal{I} = \lbrace \bm{y}, \bm{x} \rbrace$ of concatenated noisy measurements of beam responses, internal forces and applied loads $\bm{y}$ at locations $\bm{x}$. The problems we address with our model are:	 
	 \begin{itemize}
	 	\setlength\itemsep{0pt}
	 	\item[i)] to solve the inverse problem of identifying constant bending $EI$ and shear $kGA$ stiffness estimates based on collected heterogeneous and multi-fidelity data;
	 	\item[ii)] to estimate beam unobserved responses and internal forces at arbitrary locations $\bm{x}_\star$;
        \item[iii)] to find an optimal finite set of locations $\bm{x}$ where sensors are to be installed.
	 \end{itemize} 
 	To this end, in Section~\ref{sec:32} the physics-informed GP model is formulated, followed by the optimisation strategy, which includes solving the inverse problem in Section~\ref{sec:training} (objective i). Section~\ref{sec:34} defines the method for inference of unobserved responses (objective ii). Finally, Section~\ref{sec:SensorPlacement} uses the defined model for the selection of sensor locations and elaborates on the novelties of the proposed method (objective iii).	
	\subsection{Problem formulation} \label{sec:32}	\par
\begin{figure*}[h]
		\centering
		\includegraphics[width=1\textwidth]{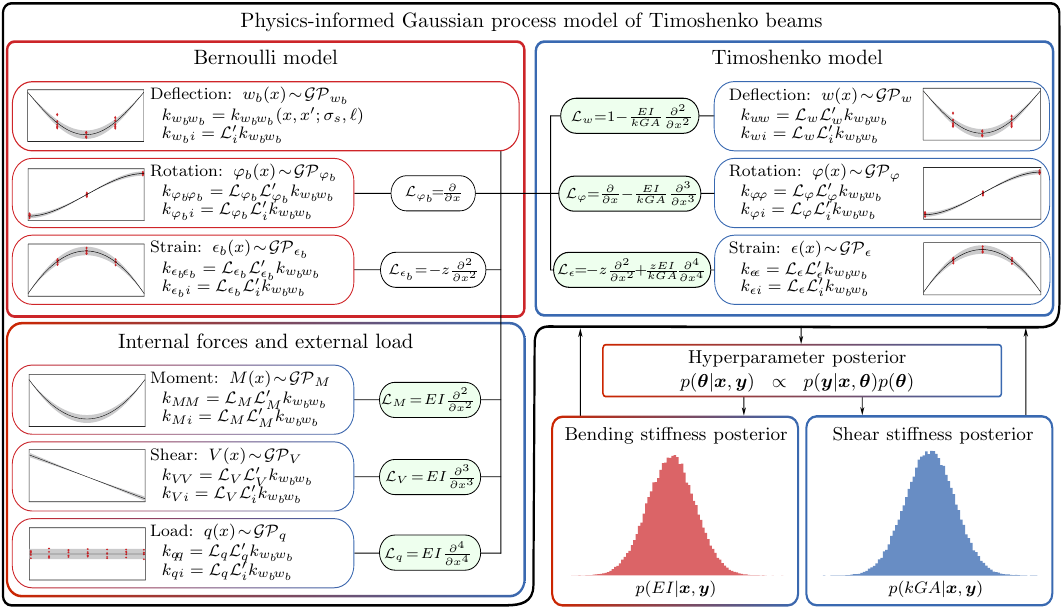}
		\caption{Physics-informed Gaussian process model of a Timoshenko beam. The models are first derived for each physical property according to the Bernoulli beam theory (red box). Combining the rotation and shear effects from the Bernoulli model yields Timoshenko's beam theory (blue box). The underlying structural stiffness $EI$ and $kGA$ are part of the hyperparameter vector $\bm{\theta}$, and can be identified from collected data $\bm{y}$ (red points) at locations $\bm{x}$ while optimizing the GP model.}\label{fig:s000_PIGPFramework}	
	\end{figure*}

	We consider at first an Euler-Bernoulli beam with constant bending stiffness subjected to a static load, derive physics-informed models for all its quantities (deflections, rotations, strains, bending moments, shear forces and applied loads), and show how they can be extended to account for shear deformations according to the Timoshenko model, assuming a constant shear stiffness. The deflection at any given position is assumed to be drawn from a zero-mean Gaussian process	
	\begin{equation}
		w_b \sim \mathcal{GP}_{w_b} \left( 0, k_{w_b w_b}\right),
		\label{eq:GPwb}
	\end{equation}	
	\noindent where $ k_{w_b w_b} = k_{w_b w_b}(x,x')$ is a covariance kernel. The zero-mean assumption in the prior of Eq.~\ref{eq:GPwb} does not mean the deflections have zero mean, which is unrealistic from a structural mechanics perspective. Rather, it implies that the predictive mean (shown later in Eqs.~\ref{eq:predFcn} and~\ref{eq:postMean}) follows no specific parametrized model, and is fully based on the training data, captured by the GP's covariance functions. Several kernel options are available in the literature, and tailoring its structure to a specific problem is a powerful way to account for prior knowledge of the function to be approximated. The Squared Exponential (SE) kernel is herein used to model $k_{w_b w_b}$ as it is continuous, smooth and infinitely differentiable~\cite{duvenaudStructureDiscoveryNonparametric2013, micchelliUniversalKernels2006}. The covariance kernel is calculated as:	
	\begin{equation}
		k_{w_b w_b}(x,x';\sigma_s,\ell) = \sigma_s^2 \mathrm{exp}  \left( -\frac{1}{2} \left( \frac{ x-x'}{\ell} \right)^2 \right),
		\label{eq:kSE}
	\end{equation}	
	\noindent where $x$ and $x'$ are spatial coordinates, $\sigma_s^2$ is a variance measure and $\ell$ controls the covariance's length scale. 	
	The applied load is in a linear relationship with the deflection response through the linear operator $\mathcal{L}_q = EI \frac{\partial^4}{\partial x^4}$ (cf. Eq. \ref{eq:Load}), and when applied to $\mathcal{GP}_{w_b}$ yields a Gaussian process for the load 	
	\begin{equation}
		q \sim \mathcal{GP}_q \left( 0, k_{qq} \right),
		\label{eq:loadGP}
	\end{equation}
	\noindent with	
	\begin{equation}
		k_{qq} = \mathcal{L}_q \mathcal{L}_q' k_{w_b w_b} = EI\frac{\partial^4}{\partial x^4} \left( EI \frac{\partial^4}{\partial x'^4} k_{w_b w_b} \right),
		\label{eq:kqq}
	\end{equation}	
	\noindent where the dependency on $\sigma_s$, $\ell$ and the spatial coordinates is omitted for simplicity. To describe the connection between the deflection and the loads, the cross-covariance kernels are calculated by:	
	\begin{gather} \label{eq:kwbq}
		\begin{split}
			&k_{q w_b} =  \mathcal{L}_q k_{w_b w_b} =EI \frac{\partial^4}{\partial x^4} k_{w_b w_b}, \\
			&k_{w_b q} =  \mathcal{L}_q' k_{w_b w_b} = EI \frac{\partial^4}{\partial x'^4} k_{w_b w_b}. \\
		\end{split}
	\end{gather}		
	\noindent In both formulations, the constant bending stiffness $EI$ is built into the covariance kernel via the differential equation and, if not known, can be identified during the training of the model. The optimisation process is discussed in Sec. \ref{sec:training}. The kernels for bending moments and shear forces are calculated similarly by applying the linear operators defined in Equations \ref{eq:TimoMoment} and \ref{eq:TimoShear} to $k_{w_b w_b}$, respectively, yielding	
	\begin{gather}
		\begin{split}
			\label{eq:kmmvv}
			&k_{M \! M} = \mathcal{L}_M \mathcal{L}_M' k_{w_b w_b} = EI \frac{\partial^2}{\partial x^2} \left( EI \frac{\partial^2}{\partial x'^2} \ k_{w_b w_b} \right), \\
			&k_{V V} = \mathcal{L}_V \mathcal{L}_V' k_{w_b w_b} = EI \frac{\partial^3}{\partial x^3} \left( EI \frac{\partial^3}{\partial x'^3} \ k_{w_b w_b} \right),
		\end{split}
	\end{gather}	
	\noindent while the remaining GP models for bending-related rotations and strains are obtained similarly. In Timoshenko's model, the total cross-section rotation is a combination of bending and shear effects. To relate the deflection due to bending with the combined (bending and shear) rotation in the cross-section, the operators in Eqs. \ref{eq:TimoShear}  and \ref{eq:TimoFullRot} are applied to $k_{w_b w_b}$ resulting in:	
	\begin{equation}
		\label{eq:kwbvarphi}
        k_{w_b \varphi} = \left( \mathcal{L}_{\varphi_b}' - \frac{\mathcal{L}_V'}{kGA} \right) k_{w_b w_b} = \left( \frac{\partial}{\partial x'} - \frac{EI}{kGA} \frac{\partial^3}{\partial x'^3} \right) k_{w_b w_b},  
	\end{equation}
    \noindent and the combined rotation kernel is obtained by
    \begin{equation}
		\label{eq:kvarphivarphi}
        k_{\varphi \varphi} = \left( \mathcal{L}_{\varphi_b} - \frac{\mathcal{L}_V}{kGA} \right) k_{w_b \varphi} =  \left( \frac{\partial}{\partial x} - \frac{EI}{kGA} \frac{\partial^3}{\partial x^3} \right) k_{w_b \varphi},
	\end{equation}
	\noindent where similarly to the bending stiffness $EI$, the constant shear stiffness $kGA$ is now a part of the covariance kernel and can be learned from data in case it is not known. The kernel for deflections on Timoshenko's model is obtained by expanding and integrating $k_{\varphi \varphi}$, which yields
    \begin{gather}
		\begin{split}
		k_{w w} =&   \iint k_{\varphi \varphi} \ \partial x \partial x' =  \left( 1 - \frac{EI}{kGA} \frac{\partial^2}{\partial x^2} - \frac{EI}{kGA} \frac{\partial^2}{\partial x'^2} \right. \\
                & \left. +  \left( \frac{EI}{kGA} \right)^2 \frac{\partial^2}{\partial x^2} \frac{\partial^2}{\partial x'^2} \right) k_{w_b w_b},
			\label{eq:kwwshear}
	   \end{split}
	\end{gather}	
	\noindent and the strain kernel in the Timoshenko model is obtained similarly by applying the linear operator defined in Eq.~\ref{eq:TimoStrain}. The remaining covariance functions that define the physics-informed machine learning model are obtained through the application of the proper differential equations, as schematically shown in Fig. \ref{fig:s000_PIGPFramework} and given in full in \ref{app:Kernels}. Equivalently as to the standard differential equation model, an Euler-Bernoulli physics-informed Gaussian process is obtained when the shear stiffness $kGA$ is sufficiently high, as the shear rotations are consequently negligible and all the shear-related terms in the covariance kernels tend to zero. 	
	Generally, the true values of responses, internal forces and applied loads cannot be directly accessed, and can either be measured with some arbitrary quality depending on the sensing equipment, or assumed with a particular uncertainty. Taking deflections as an example, a measurement $\bm{w}$ at finite locations $\bm{x} \in \mathcal{R}^N$ takes the form 	
	\begin{equation}
		\bm{w} = f(\bm{x}) + \bm{\delta}(\bm{x}),
		\label{eq:beamdef}
	\end{equation}	
	\noindent where $\bm{\delta}(\bm{x}) \sim \mathcal{N}(0, \sigma_{n,w}^2\bm{I})$ accounts for the stationary white noise in the measurement system, defined by a variance $\sigma_{n,w_b}^2$, and $\bm{I} \in \mathcal{R}^{N \times N}$ is the identity matrix. Modelling $f(\bm{x})$ as a zero-mean Gaussian process with covariance kernel calculated by Eq. \ref{eq:kwwshear}, an extended covariance matrix that accounts for uncertainties is given as:	
	\begin{equation}
		\bm{K}_{ww}^{\sigma_{n\!,\!w}} = k_{ww}(\bm{x}, \bm{x}) + \sigma_{n,w}^2\bm{I},
		\label{eq:noisycov}
	\end{equation}	
	\noindent while the other quantities follow the same logical argument. With the complete set of covariance matrices, the prior physics-informed model can be represented as a multi-output Gaussian process,	
	\begin{equation}
		\begin{bmatrix}
			\bm{w}, \bm{\varphi}, \bm{\epsilon}, \bm{M}, \bm{V}, \bm{q}
		\end{bmatrix}^{\mathrm{T}} = \mathcal{GP} \left( \bm{0}, \bm{K} \right),
	\end{equation}	
	\noindent where the covariance matrix $\bm{K}$ is calculated as	
	\begin{equation}
		\label{eq:allKernelsCombined}
		\bm{K} = 
		\begin{bmatrix}
			\bm{K}_{ww}^{\sigma_{n\!,\!w}} \! & \bm{K}_{w \varphi} \! & \bm{K}_{w \epsilon} \! & \bm{K}_{w \!M} \! & \bm{K}_{wV} \! & \bm{K}_{wq} \!\\
			\bm{K}_{\varphi w} \! & \bm{K}_{\varphi\varphi}^{\sigma_{n\!,\!\varphi}} \! & \bm{K}_{\varphi \epsilon} \! & \bm{K}_{\varphi \! M} \! & \bm{K}_{\varphi V} \! & \bm{K}_{\varphi q} \!\\
			\bm{K}_{\epsilon w} \! & \bm{K}_{\epsilon \varphi} \! & \bm{K}_{\epsilon\epsilon}^{\sigma_{n\!,\!\epsilon}} \! & \bm{K}_{\epsilon \! M} \! & \bm{K}_{\epsilon V} \! & \bm{K}_{\epsilon q} \!\\
			\bm{K}_{M \! w} \! & \bm{K}_{M \! \varphi} \! & \bm{K}_{M \! \epsilon} \! & \bm{K}_{M \!M}^{\sigma_{n\!,\!M}} \! & \bm{K}_{M \!v} \! & \bm{K}_{M \!q} \!\\
			\bm{K}_{Vw} \! & \bm{K}_{V \varphi} \! & \bm{K}_{V \epsilon} \! & \bm{K}_{V \!M} \! & \bm{K}_{VV}^{\sigma_{n\!,\!V}} \! & \bm{K}_{Vq} \!\\
			\bm{K}_{qw} \! & \bm{K}_{q \varphi} \! & \bm{K}_{q \epsilon} \! & \bm{K}_{q \! M} \! & \bm{K}_{qV} \! & \bm{K}_{qq}^{\sigma_{n\!,\!q}} \!
		\end{bmatrix}.
	\end{equation}	\par
    Structural boundary conditions are taken into account via the addition of an artificial and noise-less data set of the appropriate response. For example, in the case of a simply supported beam with length $L$, an additional set of locations $\bm{x}_w^{\mathrm{BC}} = \lbrace 0, L \rbrace$ with boundary condition information $\bm{y}_w^{\mathrm{BC}} = \lbrace 0, 0 \rbrace$ and fixed noise $\sigma_{n,w,\mathrm{BC}}^2 = 0$~m is supplied to the model along with measurement data. This enforces the predictive mean to match $\bm{y}_w^{\mathrm{BC}}$, while the predictive variance at $\bm{x}_w^{\mathrm{BC}}$ collapses to zero. Different approaches for this problem exist in the literature, for instance by deriving the covariance kernels using Green's functions or using specific basis functions for $k_{w_b w_b}$~\cite{alvarezLinearLatentForce2013a,paciorekBayesianSmoothingGaussian2007a}, and although they provide a more mathematically elegant solution to the problem, they also constrain the GP model to a particular set of structural systems. \par
	The formulation of $\bm{K}$ assumes that one dataset from each of the random fields, i.e. displacements, rotations, strains, moments, shears and loads, is available. This assumption can be relaxed to accommodate, on one hand, lack of data when no information is available, but also to account for multiple datasets on the same data type, for instance, when multiple displacement sensor sets of different quality are used to monitor a structure. The framework provides, therefore, a robust and physics-informed model for data fusion. 
	\subsection{Training and identification structural stiffness}\par
	\label{sec:training}	
	The model parameters, including the covariance kernel variables, structural stiffness and data set noises, if not known a priori, are included in a vector in $\bm{\theta} = \lbrace \sigma_s^2, \ell, EI, kGA, \sigma_w^2, \sigma_r^2, ... \rbrace$, and can be identified from the data via different optimisation schemes. In this work, a fully-Bayesian approach is adopted, and a parameter posterior distribution is defined as:	
	\begin{equation}
		p(\bm{\theta} | \bm{y}, \bm{x}) \propto p(\bm{y} | \bm{x}, \bm{\theta}) p(\bm{\theta}).
	\end{equation}
	The parameter density $p(\bm{\theta})$ can be arbitrarily defined based on prior knowledge of each of the variables in $\bm{\theta}$, including their correlations if they exist. In the particular case of flat priors, that is, $p(\theta_i) = \mathcal{U}(-\infty, \infty)$ for all $\theta_i \in \bm{\theta}$, no knowledge is assumed and the results are equivalent to a maximum likelihood estimation~\cite{cressieStatisticsSpatialData2015}. The likelihood $p(\bm{y} | \bm{x}, \bm{\theta})$ can be analytically calculated for Gaussian processes, and amounts in log form to~\cite{rasmussenGaussianProcessesMachine2006}:	
	\begin{gather}
		\begin{split}
			\mathrm{log} \ p(\bm{y} | \bm{x}, \bm{\theta}) = -\frac{1}{2} \bm{y}^\mathrm{T}\bm{K}^{-1} \bm{y} -\frac{1}{2} \mathrm{log} |\bm{K}| - \frac{n}{2} \mathrm{log} 2 \pi,	
		\end{split}
		\label{eq:maxLikEst}
	\end{gather}	
	\noindent where $|\cdot|$ denotes the determinant operation and $n$ is the number of data points available during training. Sampling from $p(\bm{\theta} | \bm{y}, \bm{x})$ is typically achieved using variational methods or techniques based on Markov chain Monte Carlo~\cite{nealMonteCarloImplementation1997, gelfandGibbsSampling2000}. In this particular study, the Metropolis-Hastings (MH) algorithm~\cite{hastingsMonteCarloSampling1970} is used to draw from the posterior distribution. The MH algorithm creates a Markov chain by iteratively sampling new candidate parameters $\bm{\theta}_*$ from a proposal distribution $g(\bm{\theta})$, and evaluating a weighted posterior probability ratio $p$. The new samples are accepted with probability $p \geqslant a \sim \mathcal{U}(0,1)$, and rejected otherwise. A burn-in is enforced to the initial $n_b$ values of the chain to isolate stable behaviour, and a thinning of $n_t$ samples is applied to ensure i.i.d. conditions. An overview of the MH algorithm is given in Alg. \ref{alg:MH}.\par
	\begin{algorithm}[h]
		\caption{Metropolis Hastings algorithm}
		\begin{algorithmic}[0]
			\State{Initialize $\bm{\theta}_0$}
			\For{$i=0,1,2,...,N-1$}
			\State{Sample $\bm{\theta}_* \sim g(\bm{\theta}_i)$}
			\State{Compute the acceptance probability}
			\State{$p = \text{min} \Bigg\{  1, \, \dfrac{p(\bm{\theta}_*|\bm{y},\bm{x}) g(\bm{\theta}_i|\bm{\theta}_*)}{p(\bm{\theta}_i|\bm{y},\bm{x}) g(\bm{\theta}_*|\bm{\theta}_i)} \Bigg\}$} 
			\State{Sample $a \sim \mathcal{U}(0,1)$} 
			\State{$\bm{\theta}_{i+1} = $ $\begin{cases}
					\bm{\theta}_*, & \text{if  } (p \geqslant  a)\\
					\bm{\theta}_i, & \text{otherwise}
				\end{cases}$}
			\EndFor
			\State{$\bm{\theta} \xleftarrow{\mathrm{burn-in, thin}} \bm{\theta}_{i\ =\ \lbrace n_b,\  n_b+n_t,\  n_b+2n_t,\ ...,\ N-1 \rbrace}$}
		\end{algorithmic}
		\label{alg:MH}
	\end{algorithm}	
	\subsection{Prediction of unobserved responses} \label{sec:34}	\par
	After identifying the optimal parameters, including the bending and shear stiffnesses, predictions on the quantity of interest $f_i (x_{\star})$ at a location $x_\star$ can be made by marginalising over the parameter posterior as  
    \begin{equation} \label{eq:fullpost}
		p(f_i (x_\star) | x_\star, \bm{y}, \bm{x}) = \int p(f_i (x_\star) | x_\star, \bm{y}, \bm{x}, \bm{\theta}) p(\bm{\theta} | \bm{y},\bm{x}) d \bm{\theta}.
	\end{equation} 
    \noindent The predictive posterior is a Gaussian process and takes the closed form
    \begin{equation}
		p({f}_i (x_{\star}) | {x}_\star, \bm{y}, \bm{x}, \bm{\theta}) = \mathcal{N} \left( \mu_\star, \sigma^2_\star \right), 
		\label{eq:predFcn}
	\end{equation}    
    \noindent where $\mu_\star$ and $\sigma_\star^2$ are the predictive mean and variance of quantity of interest $i$ at location $x_\star$, given by:	
	\begin{align}
		\mu_\star =& \bm{k}_\star^{\mathrm{T}} \bm{K}^{-1} \bm{y}, \label{eq:postMean} \\
		\sigma_\star^2 =& k_{\star\star} - \bm{k}_\star^{\mathrm{T}} \bm{K}^{-1} \bm{k}_\star. \label{eq:postStdev}
	\end{align}	
	In the previous formulation, $\bm{K}$ is the full covariance matrix of the measurement data set, including the noise parameters, as shown in Eq. \ref{eq:allKernelsCombined}, $\bm{k}_\star$ is the cross-covariance between measurements and the prediction point, calculated by:	
    \begin{gather}
    \begin{split}
        \bm{k}_\star =& [ k_{iw}(x_\star,\bm{x}_w), k_{i\varphi}(x_\star,\bm{x}_\varphi), k_{i\epsilon}(x_\star,\bm{x}_\epsilon), k_{iM}(x_\star,\bm{x}_M), \\ 
        & k_{iV}(x_\star,\bm{x}_V), k_{iq}(x_\star,\bm{x}_q) ]^{\mathrm{T}},
    \end{split}
    \end{gather}
	\noindent and $k_{\star\star} = k_{ii} (x_\star,x_\star)$ is the covariance model for the specific quantity of interest at the unobserved location.     
    The integral in Eq.~\ref{eq:fullpost} is generally intractable, and an approximate solution for the predictive model is obtained numerically through a Monte Carlo approach:	
	\begin{equation}
		p(f_i(x_\star) | x_\star, \bm{y}, \bm{x}) \approx \frac{1}{N} \sum_{n=1}^{N} p(f_i(x_\star) | x_\star, \bm{y}, \bm{x}, \bm{\theta}_n) , 
		\label{eq:predBayes}
	\end{equation}	
	\noindent where $\bm{\theta}_n \sim p(\bm{\theta} | \bm{y}, \bm{x})$ are draws from the parameter posterior distribution. Due to the assumption of Gaussian noise, the predictive posterior takes the form of a multivariate mixture of Gaussians~\cite{lalchandApproximateInferenceFully2020}, and its two first moments are given respectively by	
	\begin{align}
		\mu_\star = &  \frac{1}{N} \sum_{n=1}^{N} \mu_{\star,n},\\
		\sigma_\star^2 = &  \frac{1}{N} \sum_{n=1}^{N} \sigma_{\star,n}^2 + \frac{1}{N} \sum_{n=1}^{N} (\mu_{\star,n}-\mu_\star)^2,
	\end{align}	
	\noindent with $\mu_{\star,n}$ and $\sigma_{\star,n}$ calculated by Eqs. \ref{eq:postMean} and \ref{eq:postStdev} using the parameters $\bm{\theta}_n \sim p(\bm{\theta} | \bm{y}, \bm{x})$. \par
	\section{Physics-informed sensor placement}	\label{sec:SensorPlacement} \par	
	\subsection{Placement optimisation via entropy minimization}	\par
	The physics-informed Gaussian process derived in Sec. \ref{sec:GPModel} treats each of the physical quantities (deflections, rotations, moments, shear and loads) as probabilistic fields. Within that notion, a good sensor set $\mathcal{S}$ is one that reduces the uncertainty in the remainder of the specific domain of interest $\mathcal{D} \setminus \mathcal{S}$~\cite{krauseNearOptimalSensorPlacements2008}. Because each of the physical quantities is a GP, this uncertainty can be measured in terms of the domain entropy. Mathematically, this is calculated as
    \begin{gather}
    \begin{split}
        H& \left( x_{\mathcal{D} \setminus \mathcal{S}} | x_{\mathcal{S}} \right) = \\ 
        &- \iint p(x_{\mathcal{D} \setminus \mathcal{S}}, x_{\mathcal{S}}) \ \mathrm{log} \left( \frac{p(x_{\mathcal{D} \setminus \mathcal{S}}, x_{\mathcal{S}})}{p(x_{\mathcal{S}})} \right) d x_{\mathcal{D} \setminus \mathcal{S}} d x_{\mathcal{S}},
    \end{split}
    \end{gather}	
	\noindent where $x_{\mathcal{D} \setminus \mathcal{S}}$ and $x_{\mathcal{S}}$ are spatial coordinates belonging to the sets of unobserved and observed locations, respectively. Minimizing the entropy of the unobserved locations, in turn, corresponds to finding the set of sensor positions that are most uncertain of each other~\cite{krauseNearOptimalSensorPlacements2008}. Therefore, a set $\mathcal{S}$ containing the selected locations is obtained by the following optimisation problem:	
	\begin{equation}
		\mathcal{S}_{\mathrm{opt}} = \underset{\mathcal{S} \subset \mathcal{D}}{\mathrm{argmin}} \ H \left( {x}_{\mathcal{D} \setminus \mathcal{S}} | {x}_{\mathcal{S}} \right) = \underset{\mathcal{S} \subset \mathcal{D}}{\mathrm{argmax}} \ H \left( {x}_{\mathcal{S}} \right).
	\end{equation}	
	This type of combinatorial problem is common in many applied sciences and is proven to be NP-hard~\cite{koExactAlgorithmMaximum1995}, making its direct use impractical in most cases as the computational complexity expands in a non-polynomial manner. Consequently, a greedy approach is generally adopted~\cite{mckayComparisonThreeMethods2000,cressieStatisticsSpatialData2015,christodoulouEntropyBasedSensorPlacement2013}, where the location of maximum entropy is sought given the sensors that are currently placed, that is,	
	\begin{equation}
		x_k = \underset{x \subset \mathcal{D} \setminus \mathcal{S}}{\mathrm{argmax}} \ H \left( x_\star | \bm{x}_{\mathcal{S}} \right),
	\end{equation} 
	\noindent with $k \in \lbrace 1, ..., n_s \rbrace$ for a total of $n_s$ sensors, and the newly selected location $x_k$ is added to the set of sensor positions at each iteration. For a normally distributed random variable, as is the case for each discrete position $x$ in a GP, such entropy is calculated as:	
	\begin{equation}
		H \left( x_{\star} | \bm{x}_{\mathcal{S}} \right) = \frac{1}{2} \mathrm{ln} \left( 2 \pi \mathrm{e} \sigma_{x_\star | \bm{x}_{\mathcal{S}}}^2  \right) ,
	\end{equation}	
	\noindent where $\sigma_{x_\star | \bm{x}_{\mathcal{S}}}$ is the proposed location's standard deviation conditioned on the previously placed sensors. It is worth noting that the total uncertainty and the correspondent entropy for GPs are functions exclusively of their location $x$. Thus, sensor placement optimisation can be carried out before any data from the physical model is observed.	
	\subsection{Physics-informed placement effects}	\par
	For the physics-informed Gaussian process model from Sec. \ref{sec:GPModel}, the total variance $\sigma^2_\star$ at position $x_\star$ is calculated according to Eq. \ref{eq:postStdev}, such that the entropy is obtained as
    \begin{equation}
		H \left( x_{\star} | \bm{x}_{\mathcal{S}} \right) = \frac{1}{2} \mathrm{ln} \left( 2 \pi \mathrm{e} \left( k_{\star\star} - \bm{k}_\star^{\mathrm{T}} \bm{K}^{-1} \bm{k}_\star \right)  \right) .
	\end{equation}	
    \noindent The covariance functions that generate $\bm{k}_{\star}$, $\bm{k}_{\star \star}$ and $\bm{K}$ are derived using the underlying differential equations, yielding, in turn, a physics-informed sensor placement strategy. In addition to accounting for the physical laws governing each process, the GP model allows for the consideration of boundary conditions by including noise-less datasets for the appropriate physical quantity at specific locations. These boundary conditions will in turn affect the total entropy at position $x_\star$ even for cases where the BC and the target physical quantity are not the same. \par	
    \begin{figure*}[h]
		\centering
		\includegraphics[width=\textwidth]{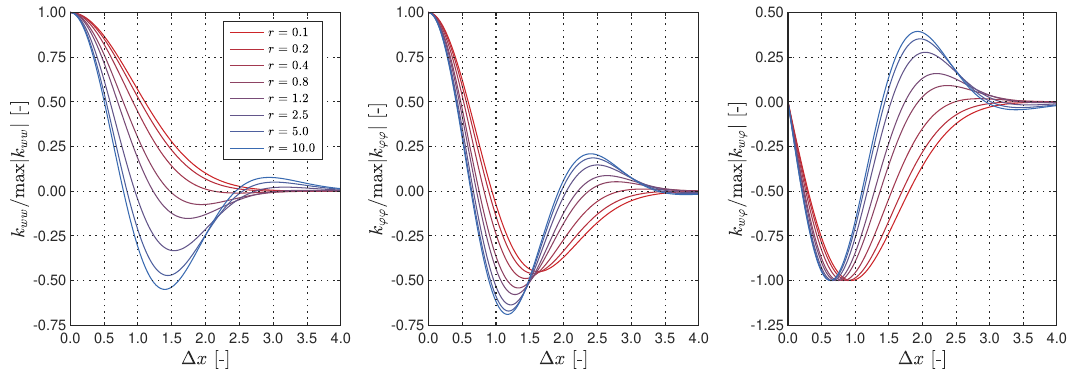}
		\caption{Influence of the rigidity $r$ on the physics-informed covariance kernels for (left to right) deflections $k_{ww}$, rotations $k_{\varphi \varphi}$, and the cross-covariance kernel $k_{w \varphi}$ between the two, as a function of the normalised distance $\Delta x = (x-x')/\ell$.}\label{fig:s003_simplySupported_SensorPlacementKernelCheck}	
	\end{figure*}
	The variance at each candidate location, and therefore also the conditional entropy, is a function of the Normalised sensor distance $\Delta x = (x-x')/\ell$ and will vary according to both the physical domain of interest and the structural stiffness. To illustrate these properties, Fig. \ref{fig:s003_simplySupported_SensorPlacementKernelCheck} shows the covariance results of the displacements $k_{ww}$, the rotations $k_{\varphi \varphi}$, and the cross-covariance $k_{w \varphi}$ between the two domains, as a function of the normalised distance $\Delta x$ and the structural rigidity $r$. Higher rigidity values progressively decrease the distance where the covariance between two sensors is zero for both $k_{ww}$ and $k_{\varphi \varphi}$, and in turn, increase the magnitude of the negative covariance that is observed after that point. The covariances approach zero as $\Delta x$ grows, indicating a very low relation between far-away sensors. \par	
	For high rigidity values, the covariances tend to oscillate between negative and positive values, reflecting a complex behaviour between spatial points. In a low rigidity range no negative covariances are observed for the deflection domain, as it is modelled purely by the Squared Exponential kernel (c.f. Eq. \ref{eq:kSE}). The results of the cross-covariance $k_{w \varphi}$ further indicate that, because of the physics-informed nature of the model, information is shared between physical domains and different types of installed sensors, effectively reducing the total entropy and affecting the placement results. 	
	\section{Numerical study}\par
\begin{figure}[h]
		\centering
		\includegraphics[width=0.45\textwidth]{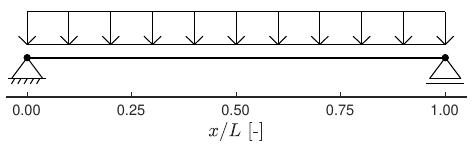}
		\caption{A simply supported beam model of length $L$, bending stiffness $EI$ and shear stiffness $kGA$, subjected to a uniform distributed load.}\label{fig:s000_SimplySuppBeam}		
	\end{figure}

	\label{sec:NumericalStudy}	
	In this section an investigation of the model derived in Sections \ref{sec:GPModel} and \ref{sec:SensorPlacement} regarding the sensor placement strategy, stiffness identification, rigidity effects and noise influence is carried out. For this purpose, a simply supported beam under uniformly distributed load (c.f. Fig.~\ref{fig:s000_SimplySuppBeam}) is taken as a fundamental example. The beam has a length $L$, and the bending stiffness $EI$ and shear stiffness $kGA$ are initially set so that $r = 1$.\par
	Following the particular structural system, the physics-informed Gaussian process model is built with the definition of the noiseless boundary conditions, such that before any measured data collection, the model's covariance matrix is defined by	
    \begin{equation*}
		\bm{K} = k_{ww}(\bm{x}_w^{\mathrm{BC}}, \bm{x}_w^{\mathrm{BC}'}),
	\end{equation*}	
	\noindent where $\bm{x}_w^{\mathrm{BC}} = \left[ 0, L \right]^T$ contains the support locations.		
	\subsection{Sensor placement optimisation}\par
	\label{sec:simpSupBeam}	
	Before any measurement data is collected, a definition of the sensor placement location set $\mathcal{S}$ must be made. Assuming that a total of $N_s = 7$ sensors shall be placed in a total of $N_p=31$ equidistant nodes distributed throughout the beam's length, the number of different placement sets is determined by the binomial coefficient	
	\begin{equation*}
		\binom{N_p}{N_s} = \frac{N_p!}{N_s! (N_p-N_s)!},
	\end{equation*}	
	\noindent and amounts to the order of $10^6$ different individual combinations of location sets.\par
\begin{figure*}[h]
		\centering
		\includegraphics[width=\textwidth]{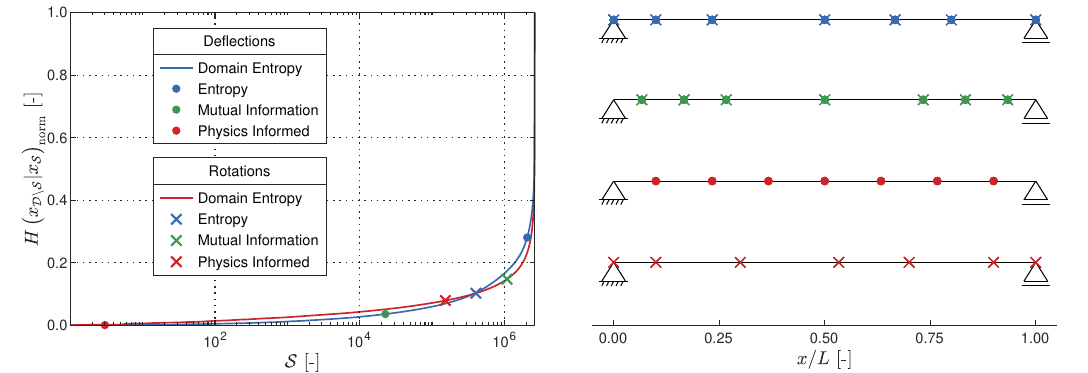}
		\caption{Left: Normalised physics-informed entropy map, for both deflection and rotation domains, for all possible combinations of $7$ sensors placed at $31$ locations, for each sensor placement criterion. Right: The correspondent sensor positions for each placement algorithm.}\label{fig:s003_simplySupported_EntropyMap}
	\end{figure*}

	
	The derived sensor placement algorithm is employed and its results are compared to the standard entropy and mutual information models using the SE kernel~\cite{cressieStatisticsSpatialData2015, shewryMaximumEntropySampling1987, krauseNearOptimalSensorPlacements2008}, as shown in Fig. \ref{fig:s003_simplySupported_EntropyMap}. The three sensor placement algorithms can efficiently determine sets of locations that minimize the domain entropy, for both deflection and rotation cases. The physics-informed model outperforms the two other algorithms, especially in the deflection domain, where a specific boundary condition is available. On the rotation domain, although no particular BC is defined, the PI model still gathers information from the cross-covariance kernel $k_{\varphi w}$, which effectively informs the model via the underlying differential equations. \par	
	The effects of the boundary conditions are also observed in the physics-informed model's placement locations. In the deflection domain, the sensors are placed away from the support locations. Similar behaviour is produced by the mutual information algorithm, which seeks the same effect by default, in contrast to the entropy minimization criterion~\cite{krauseNearOptimalSensorPlacements2008}. This might be considered beneficial, given a general-purpose sensing condition or a loosely bounded domain, but it has disadvantages for specific structural cases, e.g. a cantilever. For the rotation case of a simply supported beam, for instance, it is advantageous to have a sensor placed on top of the supports, given that rotations at those points are likely to be large. This is automatically achieved by the physics informed and the entropy models, but not by the mutual information. In addition, since no connections or distinctions across domains exist for both the entropy and the mutual information criteria, the algorithms produce the same sensor placement sets for both the deflection and rotation sets. This is evidently not the case for the physics-informed model, which draws information across different domains via the cross-covariance kernels.\par	
	\subsection{Stiffness identification}	\par
    After placing sensors and collecting measurements, the physics-informed GP model can be used to solve the inverse problem of identifying the bending stiffness $EI$ and the shear stiffness $kGA$. Throughout this section, a UDL of magnitude $q$ is applied to the structure. Analytical responses are calculated at the sensor locations and are further contaminated by white noise. The noise variance $\sigma_{n,w}^2$ for the deflection response is calculated by defining a signal-to-noise ratio	
	\begin{equation}
		\mathrm{SNR}_w = \frac{\sigma_{n,w}}{\mathrm{max}|w|},
	\end{equation}	
	\noindent while the same applies to the rotation case. In this study, we use $\mathrm{SNR}_w = \mathrm{SNR}_{\varphi} = 20$. No particular prior knowledge is assumed for the Gaussian process parameters and sensor noise standard deviations, and therefore flat prior models are used. In contrast to those, bounded uniform priors are defined for both stiffness parameters, such that	
	\begin{align}
		p(EI) =& \ \mathcal{U}(0.5, 1.5)EI_{\mathrm{true}},\\
		p(kGA) =& \ \mathcal{U}(0.5, 1.5)kGA_{\mathrm{true}},
	\end{align}	
	\noindent where $EI_{\mathrm{true}}$ and $kGA_{\mathrm{true}}$ are the numerically determined bending and shear stiffness values. The value of the applied uniformly distributed load is informed to the model at the same locations where deflection measurements are made. The model's parameters are then identified by sampling from the posterior using the Metropolis-Hastings algorithm.\par        
    \subsubsection{Sensor placement effects}\par
\begin{figure*}[h]
		\centering
		\includegraphics[width=\textwidth]{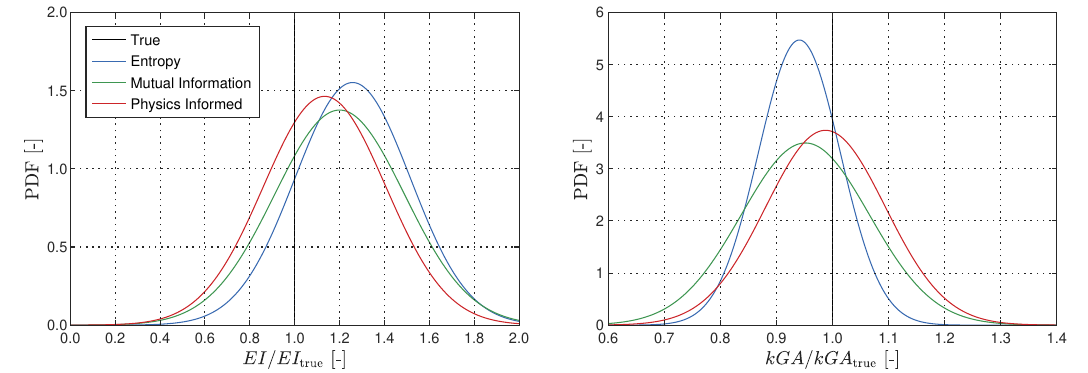}
		\caption{Probability density functions for the normalised bending stiffness $EI$ (left) and the shear stiffness $kGA$ (right) for the sensor sets obtained from the entropy, mutual information and the physics-informed criteria.}\label{fig:s003_simplySupported_SensorSetRegressionComparison}	
	\end{figure*}

    At first, we evaluate the stiffness identification results from the three optimised sensor locations. The results are shown graphically in Fig. \ref{fig:s003_simplySupported_SensorSetRegressionComparison} and in numerical form in Tab. \ref{tab:s003_simplySupportedSensorRegressionComparison}.\par	
    \begin{table}[h] 
        \centering
        \tabcolsep=0pt%
		\caption{Normalised identified posterior probability distributions for the bending $EI$ and shear $kGA$ stiffness parameters, for the different sensor placement algorithms}		
            \begin{tabular*}{0.48\textwidth}{@{\extracolsep{\fill}}lcccc}
            \toprule
            Model & \multicolumn{2}{c}{$p(EI/EI_{\mathrm{true}})$} & \multicolumn{2}{c}{$p(kGA/kGA_{\mathrm{true}})$}\\
            & $\mu$ [-] & $\sigma$ [-] & $\mu$ [-] & $\sigma$ [-] \\
			\midrule
			Entropy & 1.258 & 0.026 & 0.941 & 0.073\\
			Mutual Information & 1.200 & 0.029 & 0.952 & 0.115\\
			Physics Informed & 1.133 & 0.027 & 0.989 & 0.107\\
            \toprule
            \end{tabular*}
		\label{tab:s003_simplySupportedSensorRegressionComparison}
	\end{table}	
	Given that a moderate level of noise and a high number of sensors is installed, all three sensor sets can effectively identify the stiffness parameters. The results for the bending stiffness deviate the most from the true value, which is a consequence of the defined $r=1$. Rigidity effects will be discussed further in Sec.~\ref{sec:rigidity}. The set of sensors based on the entropy and mutual information criteria are 25.8\% and 20.0\% higher than $EI_{\mathrm{true}}$, while the physics-informed set has better accuracy, with a total error of 13.3\%. In terms of uncertainty, all three sets have a similar standard deviation. For the shear stiffness case, the model is more accurate and the identification has more similarity with $kGA_{\mathrm{true}}$, albeit they have a higher standard deviation when compared to the bending stiffness. A 5.9\% shift in the mean is obtained for the entropy criterion, while the mutual information sensor set produces a result with a 4.8\% error. The physics-informed model, however, is more accurate in the mean sense and has a value with 1.1\% error from $kGA_{\mathrm{true}}$. The mutual information and the physics-informed models have similar uncertainty in the stiffness identification, while the entropy criterion, despite having a higher deviation in the mean sense, returns a smaller uncertainty. In this manner, the influence of the sensor placement is observed in the model's parameters, and a connection can be made between the total domain entropy $H \left( x_{\mathcal{D} \setminus \mathcal{S}} | x_{\mathcal{S}} \right)$ and the posterior model $p(EI,kGA|\bm{y},\bm{x})$.	
    \subsubsection{Boundary condition effects} \label{sec:subsecBCs} \par
\begin{figure}[h]
		\centering
		\includegraphics[width=0.48\textwidth,trim=5 0 5 0,clip]{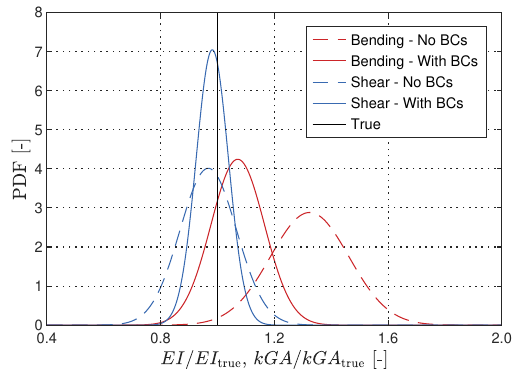}
		\caption{Estimated probability density functions for the structural normalised bending and shear stiffness. The inclusion of BCs as synthetic noise-less measurement datasets is not mandatory, but leads to a more accurate stiffness identification.}\label{fig:s001_simplySupported_ei_kga}	
	\end{figure}

    Next, we evaluate the effects of informing the model with boundary conditions on the quality of the identified stiffness values. Although in general self-evident, the importance of providing BC information is worth investigating as the physics-informed GP model does not incorporate them by default, thus maintaining generality and not conforming to one specific structural system (c.f. Sec.~\ref{sec:32}). Boundary conditions are imposed, however, by including synthetic noiseless datasets with BC information (e.g. for a simply supported beam of length $L$, $x_w^{\mathrm{BC}} = \left[ 0,L \right]^T$ and $y_w^{\mathrm{BC}} = \left[ 0,0 \right]^T$, and $\sigma_{n,w}^{\mathrm{BC}} = 0$~m), effectively conditioning the GP model to the specified points and forcing it to collapse its uncertainty at the BC locations. The lack of boundary condition information leads, as in general expected, to a less accurate and more uncertain identification of the stiffness parameters $EI$ and $kGA$, as shown in Fig.~\ref{fig:s001_simplySupported_ei_kga}. The GP model that includes BC information has increased performance in comparison to the one with no BC data sets, both in terms of the mean and the standard deviation values for $p(EI/EI_{\mathrm{true}})$ and $p(kGA/kGA_{\mathrm{true}})$. The numerical values of the identified parameters are shown in Tab. \ref{tab:s001_simplySupportedStiffnessComparison}. The effects of boundary conditions on the model predictions in unobserved responses is further discussed in Sec.~\ref{sec:predNum}.\par	
    \begin{table}[h] 
        \centering
    	\tabcolsep=0pt%
    	\caption{Normalised identified posterior probability distributions for the bending $EI$ and shear $kGA$ stiffness parameters, for the models with and without boundary condition information}		
    		\begin{tabular*}{0.48\textwidth}{@{\extracolsep{\fill}}lcccc}\toprule%
    		{Model} & \multicolumn{2}{c}{{$p(EI/EI_{\mathrm{true}})$}} & \multicolumn{2}{c}{{$p(kGA/kGA_{\mathrm{true}})$}}\\
    		& {$\mu$ [-]} & {$\sigma$ [-]} & {$\mu$ [-]} & {$\sigma$ [-]} \\
    		\midrule
    		{Without BCs} & {1.353} & {0.135} & {0.927} & {0.123}\\
    		{With BCs} & {1.079} & {0.091} & {0.984} & {0.062}\\
    		\toprule
    		\end{tabular*}
    	\label{tab:s001_simplySupportedStiffnessComparison}
    \end{table}
    \subsubsection{Noise effects}\par
    \begin{figure*}[h]
		\centering
		\includegraphics[width=\textwidth,trim=5 0 5 0,clip]{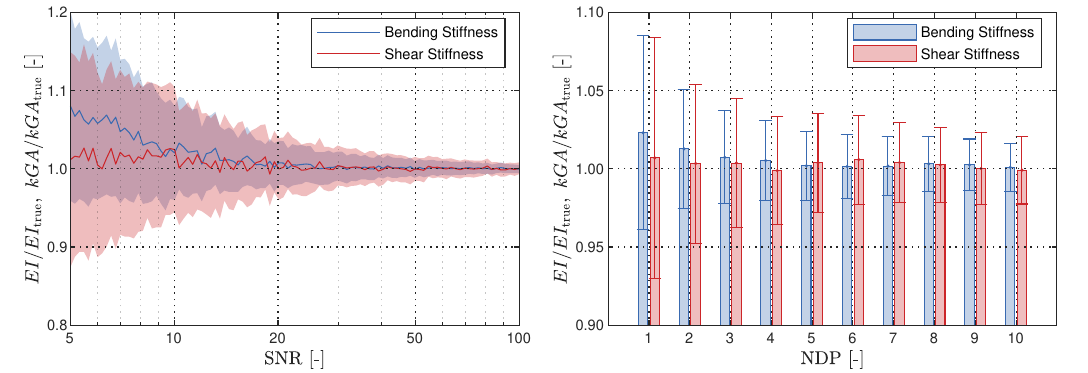}
		\caption{Bending stiffness $EI$ and shear stiffness $kGA$ identification. Left: Influence of measurement noise (mean and 95\% confidence interval) in terms of the signal-to-noise ratio (SNR). Right: Mean and 95\% confidence interval as a function of the number measurements (NDP) at each sensor location provided for training, considering a fixed $\mathrm{SNR}=10$.}\label{fig:s004_NoiseStudy}
	\end{figure*} 
    Informing the model with the appropriate boundary conditions and optimising the locations for measurement collection are essential elements for the proper identification of structural parameters. Nevertheless, the accuracy of the results is directly influenced by the quality of the measurement data. To evaluate the noise effects on the identified stiffness parameters, the model with BCs from Sec.~\ref{sec:subsecBCs} is used in this section. Analytical measurements are contaminated with white noise defined by SNRs varying from 5 to 100. To account for stochastic effects, a Monte Carlo analysis with $N_{\! M \! C} = 1000$ simulations are carried out and the outputs, in terms of the mean and standard deviation of the learned $p(EI, kGA| \bm{y}, \bm{x})$, are shown in Fig. \ref{fig:s004_NoiseStudy} (left). Results indicate that, for high amounts of noise, a slight deviation from the mean value for the shear stiffness, and a more pronounced deviation in the bending stiffness case. As the noise level decays, both structural parameters stabilize their means around the analytical values, and the model's uncertainty on the parameters decreases due to the provided measurements being less scattered around the analytical solution.\par	
	Improving the quality of the measuring devices enhances the quality of the stiffness identification. Nevertheless, this is not always a viable strategy from a technical or financial perspective. An alternative option proposed herein is to provide the model with additional measurements at each of the observed locations. This approach drastically increases the size of the data set and must be used with care as the computational complexity of the GP model scales with $\mathcal{O}(N^3)$~\cite{rasmussenGaussianProcessesMachine2006}. Although alternatives to this issue exist in the literature, e.g. partitioning the data into several groups and optimizing them individually~\cite{snelsonLocalGlobalSparse2007}, approximating the covariance matrix via sparsely sampled points~\cite{quinonero-candelaUnifyingViewSparse2005}, or using stochastic variational inference methods~\cite{hensmanGaussianProcessesBig2013}, the standard GP regression using all the data points is still considered here. To evaluate the effects of the possibly large number of data points ($\mathrm{NDP}$) provided to the model, a constant $\mathrm{SNR}=10$ is assumed, and different numbers of measurements from deflections and rotations at the physics-informed sensor location data set are added to the model. The results are shown in Fig. \ref{fig:s004_NoiseStudy} (right). The additionally provided data tends to stabilize the mean value for both bending and shear stiffness, while simultaneously reducing the uncertainty of the parameters. Nevertheless, an asymptotic limit on the standard deviation seems to exist, which is directly related to the quality of the measurement data. This becomes evident from the relation between the two plots in Fig. \ref{fig:s004_NoiseStudy}, as $\sigma_{EI}$ and $\sigma_{kGA}$ reduce with the increase of $\mathrm{NDP}$, but do not reach levels as low as the ones observed for high values of $\mathrm{SNR}$ in Fig \ref{fig:s004_NoiseStudy} (left).\par	
    \subsubsection{Rigidity effects} \label{sec:rigidity}\par
    \begin{figure*}[h]
		\centering
		\includegraphics[width=\textwidth,trim=5 0 5 0,clip]{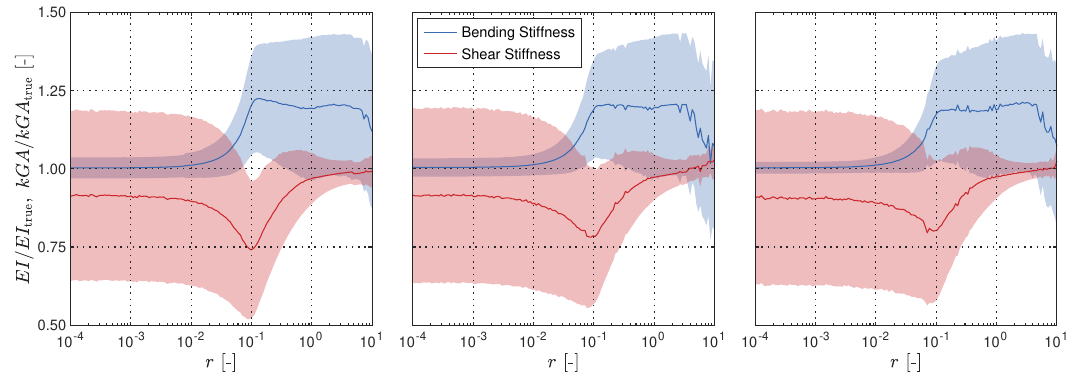}
		\caption{Mean and 95\% confidence interval of the identified bending stiffness $EI$ and shear stiffness $kGA$ in relation to the beam rigidity $r=3EI/L^2kGA$. From left to right: Regression using only deflection data, only rotation data, and both simultaneously.}\label{fig:s002_RigidityStudy}	
	\end{figure*}	
	The results discussed so far consider a fixed rigidity $r=1$. The beam response, however, is a combination of bending and shear components and is therefore directly dependent on the rigidity parameter (see Fig.~\ref{fig:FigBeam}). Due to the varying response contribution as a function of $r$, it is expected that the identified stiffness models $p(EI, kGA|\bm{y},\bm{x})$ will also vary w.r.t the rigidity. Furthermore, the previous models were all identified using data sets from deflections and rotations simultaneously. In Fig. \ref{fig:s002_RigidityStudy}, results are shown for models trained with deflections only, rotations only, or both simultaneously, considering a noise level of $\mathrm{SNR} = 10$. In all cases, the accuracy of the bending stiffness identification is increased for low rigidity levels ($r < 10^{-2}$), reflecting the physical interpretation that, at this range, the response is governed by bending effects. For rigidity values between $10^{-2}$ and $10^{0}$, neither the bending nor the shear stiffness is reliably identified, as the deflection is a combination of both stiffness values and the solution for the deflection-only GP model is not unique. Conversely, for cases where $r > 10^{0}$, the shear stiffness accuracy increases progressively, while the bending stiffness results are unreliable. \par  
	When only deflection data is provided during training (Fig. \ref{fig:s002_RigidityStudy}, left), a higher standard deviation is observed for stable rigidity ranges of $p(EI, kGA|\bm{y},\bm{x})$, in both stiffness cases. The provision of rotations reduces the stiffness uncertainty (Fig. \ref{fig:s002_RigidityStudy}, centre), as rotations are in general more sensitive to changes in model parameters~\cite{rauAssessmentFrameworkSensorbased2017}. Combining the two datasets yields, however, the best prediction quality as the physics-informed model correlated the different readings based on the full covariance matrix as shown in Eq. \ref{eq:allKernelsCombined}.\par
    \subsection{Prediction of unobserved responses} \label{sec:predNum}	\par
    The physics-informed GP model is derived for all the quantities of interest related via the differential equation. Even though available data may be limited for certain locations and specific physical responses, the nature of the GP model allows for predictions of all physical quantities. Using the sensor locations obtained via the physics-informed criterion, a comparison is now made for predictions between models learned with and without the presence of BCs. For that purpose, the simply supported beam model under UDL is defined with $r=1$ and  $\mathrm{SNR}=20$. \par    
    \begin{figure*}[h]
		\centering
		\includegraphics[width=\textwidth,trim=5 0 5 0,clip]{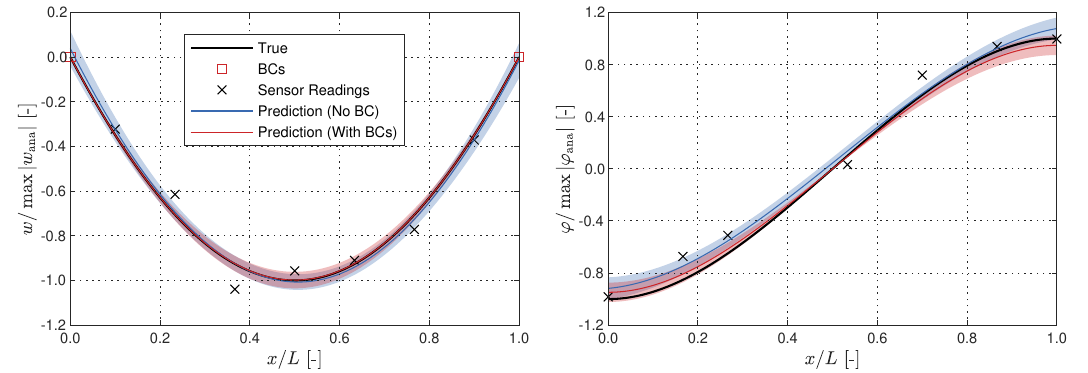}
		\caption{Simply supported beam with uniform loading: normalised deflection $w$ (left) and rotation $\varphi$ (right). The shaded area represents the 95\% confidence interval. The boundary condition on displacement reduces uncertainty at the BCs and improves prediction quality.}\label{fig:s001_simplySupported_w_phi}	
	\end{figure*} 	
    The noisy measurements for deflections and rotations, along with the respective predictions for the cases with and without BCs are shown in Fig. \ref{fig:s001_simplySupported_w_phi}. The inclusion of a noiseless boundary condition improves the quality of the predictions. The deflection uncertainty for the BC model reduces at locations closer to the supports and takes a maximum value at mid-span. In contrast, the model without BCs is characterized by a higher standard deviation throughout the length of the beam model. In addition, the prediction mean matches closely the analytical results in the case where BCs are present, while the model without BCs shows discrepancies for the mid-span prediction, and does not return a zero deflection at the supports. In the rotation case, both models stray from the analytical results. Nevertheless, the mean value of the model with BCs better approximates the true results, in comparison to the mean prediction of the model without BCs.\par	
    \begin{figure*}[h]
		\centering
		\includegraphics[width=\textwidth,trim=5 0 5 0,clip]{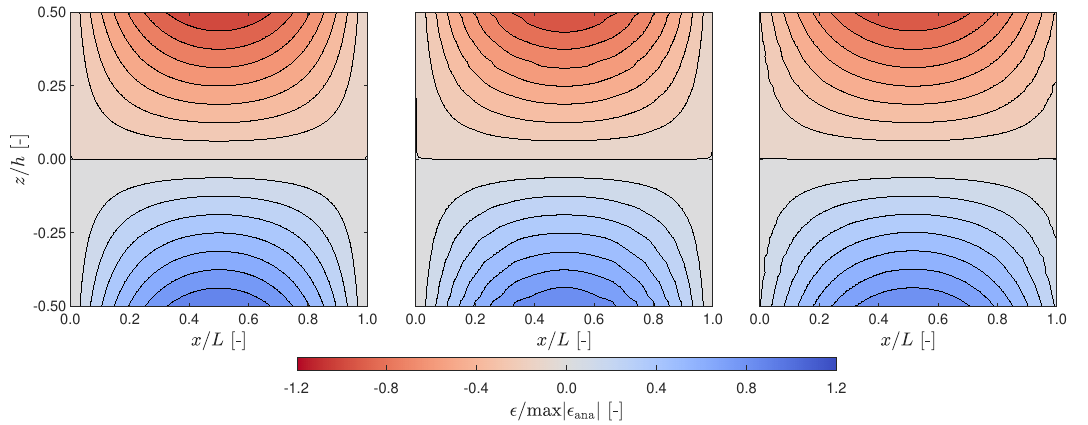}
		\caption{Simply supported beam with uniform loading: Normalised mean strain field as a function of length and section height for (left) the analytical model, (centre) the GP model with the inclusion of boundary conditions and (right) the GP model without BCs.}\label{fig:s001_simplySupported_Strain}	
	\end{figure*}
    Although no data from strain measurements and internal forces are included during the training of the physics-informed GP model, the full covariance formulation given in Eq. \ref{eq:allKernelsCombined} allows for a joint, multi-output prediction relating unobserved physical responses with the measurement data. Given the beam height $h$ and a relative distance $z$ from the neutral axis, the strain field is determined from the trained Gaussian processes models and compared to the analytical solution, as shown in Fig. \ref{fig:s001_simplySupported_Strain}. The model with BCs closely resembles the analytic solution for the strain field, especially at the cross-section extremes ($z/h = \pm 0.50$). The model without boundary condition information, however, can approximate the correct strains around the neutral axis level ($z/h=0$) but loses accuracy at the section borders, particularly close to the support locations. \par	
    \begin{figure*}[h]
		\centering
		\includegraphics[width=\textwidth,trim=5 0 5 0,clip]{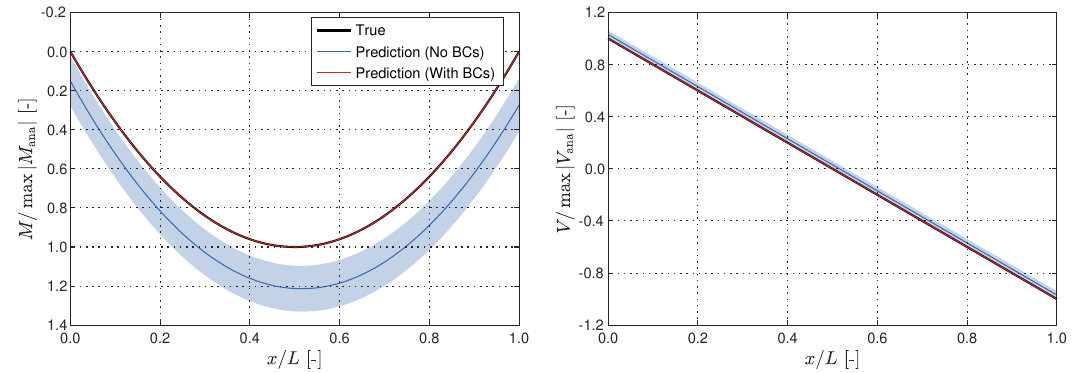}
		\caption{Simply supported beam with uniform loading: Normalised bending moment $M$ (left) and shear $V$ (right). The shaded area represents the 95\% confidence interval. The inclusion of boundary conditions decreases internal force uncertainty and increases their accuracy.}\label{fig:s001_simplySupported_M_V}	
	\end{figure*}
	The results for the internal forces are shown in Fig. \ref{fig:s001_simplySupported_M_V}. A significant bias is observed in the bending moments of the model with no BCs, along with substantial uncertainty in the predictions. Although less apparent, a deviation is also observed in the shear predictions. The model containing the information of the BCs, however, displays a virtually perfect result in terms of bending moments and shears, approximating with high precision and accuracy the two physical quantities, while their standard deviation collapses to a value substantially smaller than the non-informed model. For a better, numerical comparison of the prediction means, the root mean squared error (RMSE) for the different models is shown in Tab. \ref{tab:s001_simplySupportedBeam_RMSEComparison}.\par	 
    \begin{table*}[h] 
        \tabcolsep=0pt%
		\caption{RMSE comparison for the predictions from the models with and without boundary conditions, for all physical quantities}	
            \begin{tabular*}{\textwidth}{@{\extracolsep{\fill}}lccccc}\toprule%
            {Model} & {$w/\mathrm{max}|w_{\mathrm{ana}}|$} & {$\varphi/\mathrm{max}|\varphi_{\mathrm{ana}}|$} &
			{$\epsilon/\mathrm{max}|\epsilon_{\mathrm{ana}}|$} & {$M/\mathrm{max}|M_{\mathrm{ana}}|$} & {$V/\mathrm{max}|V_{\mathrm{ana}}|$} \\
			& {[-]} & {[-]} & {[-]} & {[-]} & {[-]} \\
			\midrule
			{Without BCs} & {$1.76 \smallcdot 10^{-2}$} & {$6.65 \smallcdot 10^{-2}$} & {$4.51 \smallcdot 10^{-2}$} & {$2.16 \smallcdot 10^{-1}$} & {$3.08 \smallcdot 10^{-2}$}\\
			{With BCs} & {$1.14 \smallcdot 10^{-3}$} & {$3.65 \smallcdot 10^{-2}$} & {$2.55 \smallcdot 10^{-2}$} & {$1.93 \smallcdot 10^{-7}$} & {$5.13 \smallcdot 10^{-7}$}\\
            \toprule
            \end{tabular*}
        \label{tab:s001_simplySupportedBeam_RMSEComparison}
	\end{table*}

	\section{Experiments}\par
	\label{sec:ExperimentalResults}
    \subsection{Setup}	\par
\begin{figure*}[h]
		\centering
		\includegraphics[width=\textwidth,trim=5 0 5 0,clip]{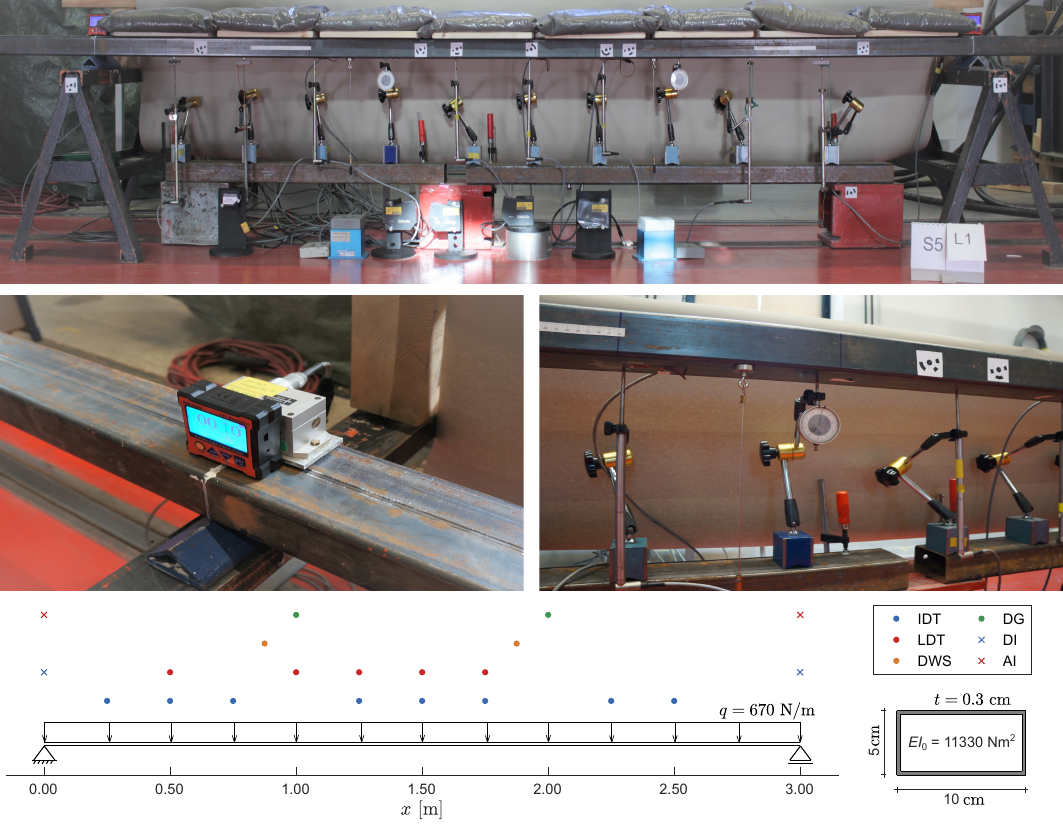}
		\caption{Top: deflection and rotation sensors positioned along the length of the beam, subjected to a uniformly distributed load. Centre: the analogue (AI) and digital (DI) inclination sensors (left) at one of the supports and (right) a detail of three deflection sensors: a displacement transducer (IDT), a draw-wire sensor (DWS) and a dial gauge (DG), while the laser displacement transducers (LDT) were installed at the floor. Bottom: a sketch of the positions of all installed sensors (left) and the beam's cross-section dimensions and thickness $t$ (right).}\label{fig:s006_Experim}	
	\end{figure*}

	We now present a validation of the constructed GP framework based on an experimental setup. The test structure consisted of a simply supported beam of $3$~m in length, as shown in Fig.~\ref{fig:s006_Experim}~(top). The beam is defined by a rectangular hollowed cross-section (c.f. dimensions in Fig.~\ref{fig:s006_Experim}, bottom right). The structural rigidity is $r=6 \cdot 10^{-4}$ and the deflections are therefore primarily governed by bending effects. The assumed bending stiffness, calculated with the section and material properties, amounts to $EI_\mathrm{0} = 11330$ Nm$^2$. The structure is subjected to a uniformly distributed load of magnitude $q = 670$ N/m, and the corresponding response, in terms of deflections, rotations and strains, was measured at various points throughout the length of the model. \par
    \begin{table}[h] 
    \centering
        \tabcolsep=0pt%
		\caption{Sensor types and properties. The DAQ has a resolution of 24 bits.}	
            \begin{tabular*}{0.48\textwidth}{@{\extracolsep{\fill}}lcc@{}}\toprule%
            {Sensor set} & {Range} & {Resolution}\\
			\midrule
			{LDT} & {400 $\pm$ 100 mm} & {DAQ} \\
            {IDT} & {0 - 50 mm} & {DAQ}\\
            {DWS} & {0 - 2000 mm} & {DAQ} \\
            {DG} & {0 - 10 mm} & {0.01 mm}\\
            {DI} & {$\pm$ 3$^{\circ}$} & {DAQ} \\
            {AI} & {$\pm$ 180 $^{\circ}$} &  {0.01 $^{\circ}$} \\
            \midrule
            {SG}  & \multicolumn{2}{c}{{Circuit: Wheatstone half-bridge, 350 $\Omega$}}\\
            \toprule
            \end{tabular*}
        \label{tab:sensorTypes}
	\end{table}	
    The deflections were measured by various types of sensors, namely five laser displacement transducers (LDT), eight inductive displacement transducers (IDT) and two draw wire sensors (DWS). The measurements were obtained using a data acquisition (DAQ) system  with 24-bit analogue-to-digital (AD) conversion. 
    In addition, analogue deflection readings were obtained using two dial gauges (DG), as shown in detail in Fig.~\ref{fig:s006_Experim} (centre right). The rotation response was measured exclusively at the support locations (c.f. Fig.~\ref{fig:s006_Experim}, centre left), simultaneously by a set of digital inclinometers (DI) on either side of the beam connected to the DAQ system and another set of analogue inclinometers (AI) at the same positions. Furthermore, strain measurements were taken at three locations at the bottom side of the structure using a Wheatstone half-bridge strain gauge (SG) circuit. Employing the DAQ system, data was acquired with a constant sampling rate of $20$ Hz. Analogue readings (dial gauge and inclinometer) were taken once the reading stabilized in a single value after the loading was applied. Details of the employed sensors are given in Table~\ref{tab:sensorTypes}, and the location of each installed sensor is indicated in Fig.~\ref{fig:s006_Experim} (bottom left).\par
\begin{figure*}[h]
		\centering
		\includegraphics[width=\textwidth,trim=5 0 5 0,clip]{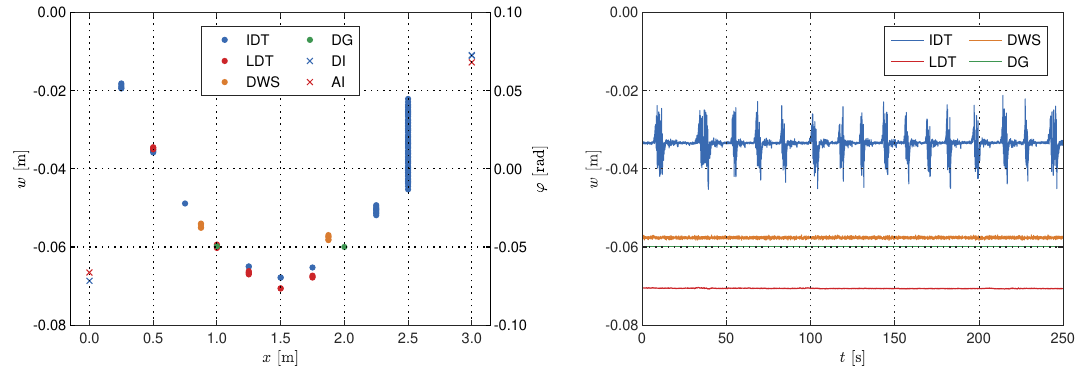}
		\caption{Left: Deflection measurements from the laser displacement transducer (LDT), displacement transducer (IDT), draw wire sensor (DWS), dial gauge (DG) sensors, along with the digital (DI) and analogue (AI) inclinometer measurements. Right: The corresponding deflection measurement time series for $x_{\mathrm{IDT}} = 2.50$ m, $x_{\mathrm{LDT}} = 1.50$ m, $x_{\mathrm{DWS}} = 1.88$ m and $x_{\mathrm{DG}} = 1.00$ m.}\label{fig:s006_SensorDataPlot}	
	\end{figure*}

	A sample of the readings from each sensor is shown in Fig. \ref{fig:s006_SensorDataPlot} (left). Analysing at first the deflection readings, the inductive displacement transducer located at $x_{\mathrm{IDT}}=2.50$ m malfunctioned during the experiment. Its time-series response, shown in Fig. \ref{fig:s006_SensorDataPlot} (right), displays an impulse-like excitation which did not take place during the static test of the structure and is not observed in the remaining sensors. In addition, the draw wire transducer at $x_{\mathrm{DWS}}=1.88$ m has a biased response, characterized by a constant deviation from the qualitative expected reading, when compared to the other sensors in its proximity. Lastly, the readings from the three laser sensors close to the midspan are higher in magnitude when compared to the readings from the IDTs installed at the same locations. The readings from the digital inclinometers are very consistent and have a low amount of noise. When compared to the analogue rotation readings, they are higher in magnitude at both support points.\par	
    \subsection{Stiffness identification}\label{sec:experimStiffID}\par
     \begin{figure*}[h]
		\centering
		\includegraphics[width=\textwidth,trim=0 0 0 0,clip]{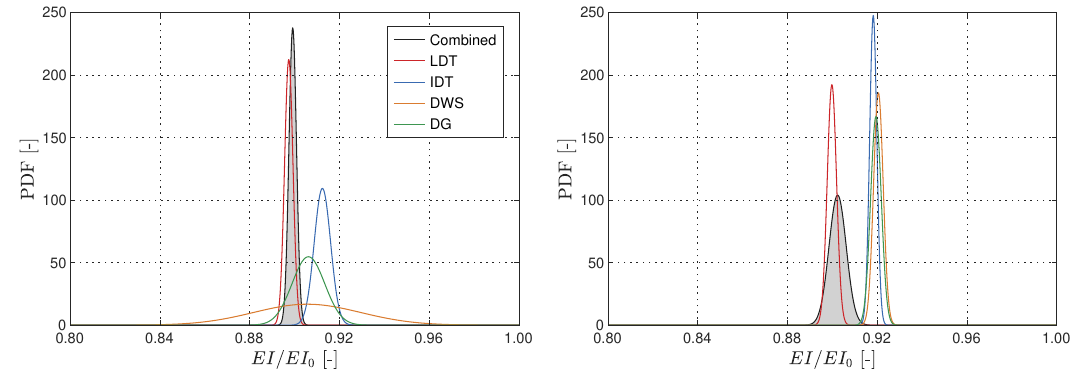}
		\caption{Normalised identified stiffness distributions for GP models without (left) and with (right) information of the rotation values measured at the supports. Models are trained separately with the laser displacement transducer (LDT), displacement transducer (IDT), draw wire sensor (DWS), dial gauge (DG) sensors, or all combined.}\label{fig:s010_sensorComparison}
	\end{figure*}	
    \begin{table*}[h] 
        \tabcolsep=0pt%
		\caption{Mean and standard deviation from the bending stiffness distributions $p(EI/EI_{0})$ identified from GPs with single sensor data sets, and the model with all data sets combined}		
            \begin{tabular*}{\textwidth}{@{\extracolsep{\fill}}lcccc@{}}\toprule%
            {Sensor set} & \multicolumn{2}{c}{{No rotations}} & \multicolumn{2}{c}{{With rotations}}\\
			& {$\mu$ [-]} & {$\sigma$ [-]} & {$\mu$ [-]} & {$\sigma$ [-]} \\
			\midrule
			{Combined} & {0.899} & {$1.7 \cdot 10^{-3}$} & {0.902} & {$3.8 \cdot 10^{-3}$}\\
			{Laser displacement transducer (LDT)} & {0.897} & {$1.9 \cdot 10^{-3}$} & {0.899} & {$2.1 \cdot 10^{-3}$}\\
		      {Inductive displacement transducer (IDT)} & {0.912} & {$3.6 \cdot 10^{-3}$} & {0.918} & {$1.6 \cdot 10^{-3}$}\\
			{Draw wire sensor (DWS)} & {0.906} & {$23.2 \cdot 10^{-3}$} & {0.920} & {$2.1 \cdot 10^{-3}$}\\
			{Dial gauge (DG)} & {0.906} & {$7.3 \cdot 10^{-3}$} & {0.919} & {$2.4 \cdot 10^{-3}$}\\
            \toprule
            \end{tabular*}
        \label{tab:s007_experiment}
	\end{table*}
	At first, we consider different Gaussian process models with $\mathrm{NDP}=7$ measurement values at each sensor location and train them with data from one individual set of deflection sensors at a time, without the inclusion of rotation readings. All the models return results around $10$\% smaller than the analytically estimated $EI_0$ (c.f. Fig.~\ref{fig:s010_sensorComparison} (left) and Tab.~\ref{tab:s007_experiment}). The model trained using the laser displacement transducer has the smallest variance, due to the low noise level in the readings. The results based on the inductive displacement transducer have a higher mean compared to the laser sensor, which is in agreement with the smaller deflections measured at midspan. In addition, a higher uncertainty is observed as a result of the faulty sensor at $x_{\mathrm{IDT}}=2.50$ m. The dial gauge results have a mean value between the LDT and IDT sensor sets but display a higher uncertainty due to the limited number of installed sensors. Finally, the draw wire sensors have a similar mean value but a much bigger variance, since only two sensors are available and one of them, at $x_{\mathrm{DWS}}=1.88$, has biased readings.\par
    To demonstrate the physics-informed GP ability of multi-fidelity sensor fusion, we now train a combined model containing the four different deflection data sets from each sensor type. To this end, a different noise parameter $\sigma_n$ is assumed for each individual set of measurements, to account for differences in sensor quality. The combined stiffness model is shown in Fig. \ref{fig:s010_sensorComparison} (left). Results are guided by the model based on the laser sensor set, as it contained the highest number of sensors and the highest SNR. Nevertheless, the mean value of the combined model is shifted towards a higher stiffness, to account for the remaining measurements, especially at midspan.\par
	Incorporating rotation readings from both the analogue and digital sensor sets in the GP model implies heterogeneous sensor fusion and alters the characteristics of the identified stiffness (c.f. Fig.~\ref{fig:s010_sensorComparison} (right) and Tab.~\ref{tab:s007_experiment}). In comparison to the pure deflection case, the quality of the identified stiffness model is increased, as rotations are generally more sensitive to changes in model parameters~\cite {rauAssessmentFrameworkSensorbased2017}. We first analyse the combination of individual deflection sensor sets with rotations. The laser sensor results have only a slight shift in mean, indicating a good agreement between the heterogeneous measurements. In all the other models, however, the mean values stabilize at a higher stiffness, reflecting the higher deflections measured by the laser displacement sensor. Notably, the variance of $p(EI/EI_0)$ for the draw-wire sensor set decreases by a factor of 11, which demonstrates how the inclusion of heterogeneous measurements adjusts the physics-informed GP model. In the case where all the measurements (deflections and rotations) are combined in a single GP model, a slight shift in the mean value of $p(EI/EI_0)$ is observed, along with an increase in variance. This may be the effect of the combination of heterogeneous measurements during the training process, as the GP model converges to an average stiffness that explains best the training data and accounts for measurement differences via an increase in uncertainty.\par	
    \subsection{Predictions of unobserved responses}\par
\begin{figure*}[h]
		\centering
		\includegraphics[width=\textwidth,trim=5 0 5 0,clip]{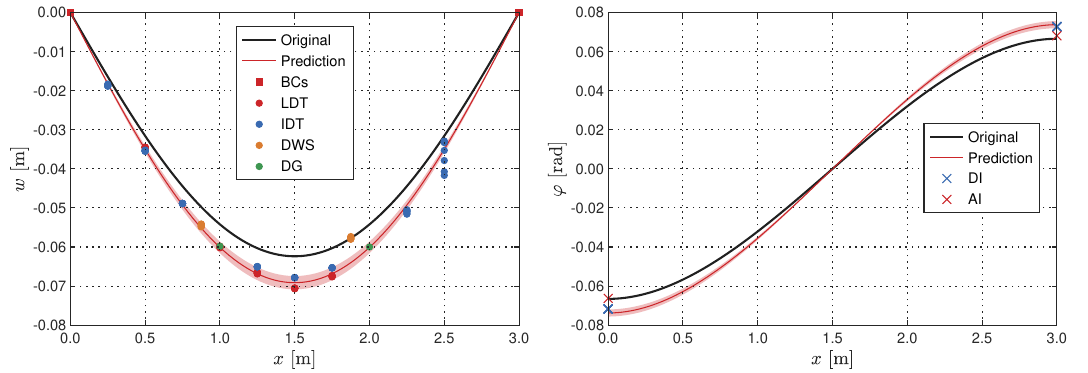}
		\caption{Inference in the physics-informed Gaussian process using the experimental data set. Left: deflection predictions, along with boundary condition information and readings from the laser displacement transducer (LDT), displacement transducer (IDT), draw wire sensor (DWS), and dial gauge (DG) sensors (left). and rotation predictions with the corresponding measurements (right). Right: rotation predictions with the training data from the digital (DI) and analogue (AI) inclinometers.}\label{fig:s007_simplySupported_w_phi}	
	\end{figure*}

    With the probability distributions for stiffness in hand, we can further use the model for predictions of physical quantities. To this end, the model for $p(EI)$ considering the combined case of all sensors from Sec.~\ref{sec:experimStiffID} is employed. At first, we focus on the deflection and rotation predictions (c.f. Fig.~\ref{fig:s007_simplySupported_w_phi}), for which data at different positions was used during training. Because the identified stiffness is lower than $EI_0$, an increase in response amplitude is observed for both deflections and rotations, when compared to the analytical model using the original assumption. The deflection predictions fit the training points accurately, and at midspan, they fall between the readings of the LDT and the IDT, which is in agreement with the intermediate average value of $EI$ (see Fig.~\ref{fig:s010_sensorComparison}, right). Prediction uncertainty decreases towards the supports, due to the effect of the noise-less boundary conditions supplied to the model during training. The noisy readings from the IDT at $x_{\mathrm{IDT}}=2.50$ m and the bias on the DWT measurements at $x_{\mathrm{DWS}} = 1.88$ m are ignored by the model, and do not influence the predictions at their respective locations.\par
    \begin{figure*}[h]
		\centering
		\includegraphics[width=\textwidth,trim=5 0 5 0,clip]{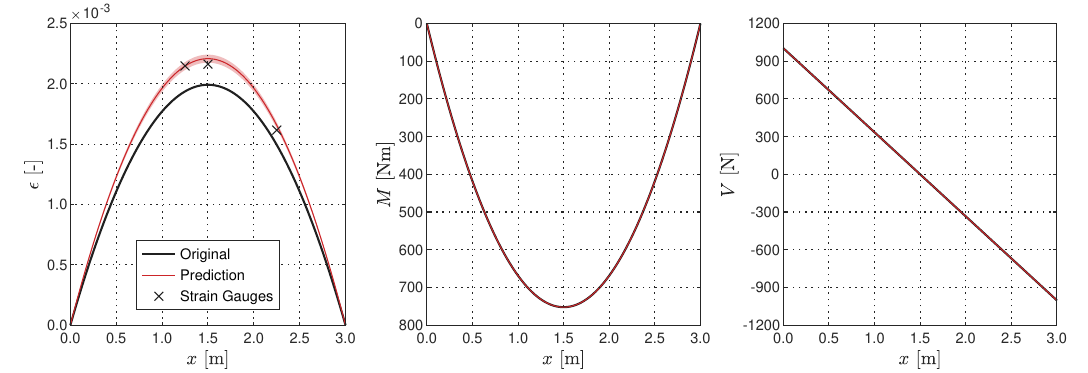}
		\caption{Prediction of unobserved responses in the physics-informed Gaussian process using the experimental data set: Strains at the bottom fibre with three strain gauge measurements used for prediction only (left), bending moments (centre) and shear forces (right). The shaded area represents the 95\% confidence interval.}\label{fig:s007_simplySupported_e_M_V}	
	\end{figure*}	
    The cross-covariances of the physics-informed GP allow predictions for physical quantities that were not included as part of the training data. In this validation case, the unobserved responses amount to strains, bending moments and shear forces. In the case of strains (c.f. Fig.~\ref{fig:s007_simplySupported_e_M_V}, left), the predicted magnitude is higher than the original model, with assumed $EI_0$. Similar to the deflections and rotations, this is explained by the lower stiffness value identified based on the GP learning. The three strain gauges installed at the bottom fibre of the cross-section are now used for the validation of the predictions. The measured data are in good agreement with the GP's strain values, falling within the 95\% confidence interval for all three locations. For the internal forces, a nearly identical result is obtained between the original and the predicted bending moments and shear forces. Because the structure is statically determined, the internal forces do not depend on the structural stiffness and are solely determined by the applied load, which is reflected by the GP model outputs. Since the load values are analytically calculated and assumed noise-less, the uncertainty in internal forces follows the same pattern and their variance is insignificant in comparison to their magnitude.\par    
	\section{Conclusions}\label{sec:conclusion}	\par
	This paper presented a physics-informed Gaussian process model for Timoshenko beam elements and its application on stiffness identification and the probabilistic prediction of unobserved responses. The developed model is able to learn from heterogeneous data types, such as deflections, rotations and strains, building correlation between them via analytically-derived cross-covariance kernels. The model seamlessly incorporates multi-fidelity measurements, aggregating data of various quality levels to reach an optimal training state. In addition, an entropy-based method for sensor placement optimisation was extended to account for the physics-informed aspect of the GP model, allowing for heterogeneous sensor placement and identifying information gains from measurement points across physical domains. \par	
    The identification of stiffness was observed to depend on the quality of the sensor placement. Furthermore, it is also connected to the structural rigidity, which was built into the GP model by construction, and the accuracy of the prediction was correlated with the stiffness component that governs the response.  It is worth noting that a basic assumption is that the collected data originates from a process described by the target PDE, in this case, the Timoshenko static beam theory, and deviations from it, e.g. cases where normal forces create additional deflections, may lead to poor stiffness identification results. In addition, due to the modelling assumptions, non-linear effects on the beam's static responses are not accounted for and are a topic for future research. The identified stiffness was subsequently used to obtain probabilistic predictions of unobserved responses, such as internal forces or deflections and rotations where sensors were not installed. This allowed for full state estimation and can be applied to locations where placing a sensor is not viable. The current strategy is limited to smooth and continuous responses, however, due to the intrinsic assumptions of the squared exponential kernel.\par	 
	The probabilistic aspect of the GP model yielded a physics-informed sensor placement optimisation method that surpassed conventional entropy-based approaches due to its ability to integrate prior structural knowledge, such as boundary conditions. This was accomplished through the inclusion of noise-free synthetic data points placed strategically within the structure. The proposed method is particularly effective for optimizing heterogeneous sensors, due to the use of cross-covariance kernels that describe the correlation of different sensor types, effectively carrying information from sensors placed across different physical domains. Results demonstrated that the proposed approach outperforms standard techniques found in the literature.\par		
	Validation of the developed model was given in the form of an experimental static test. The Gaussian process model was highly effective at identifying outliers in the form of noisy sensor readings and was capable of fusing data from heterogeneous multi-fidelity sensor types, leading to optimized stiffness values that are in closer agreement with the measured data. Nevertheless, for practical applications, the amount of measured data remains a challenge, as it demands significant computational resources.\par	
	The model presented in this paper has potential applications in the field of Structural Health Monitoring (SHM) for both mechanical and structural systems. It is particularly useful in cases involving deep and rigid beams, where traditional system identification techniques may yield poor results. In addition, the stochastic aspect of the identified parameters and the fully Bayesian response predictions provide results with valuable confidence intervals, instead of point estimates as is usually the case with traditional methods. Future studies may extend the model to more complex structural systems, opening up new avenues of research.\par
    \section*{Acknowledgements}\par
    IK gratefully acknowledges the support by the German Research Foundation (DFG) [Project No. 491258960], Darwin College and the Department of Engineering, University of Cambridge. 

	\bibliographystyle{arxiv} 
	\bibliography{references}

	\appendix
	\section{Covariance kernels}
	\label{app:Kernels}
	The derivation of all physics-informed GP models is now presented. The deflections of an Euler-Bernoulli beam are assumed to be drawn from a zero-mean Gaussian process $w_b \sim \mathcal{GP} \left( 0, k_{w_b w_b}(x,x') \right)$, where the covariance kernel is defined (c.f. Eq. \ref{eq:kSE}) by
	\begin{equation*}
		k_{w_b w_b} = \sigma_s^2 \mathrm{exp}  \left( -\frac{1}{2} \left( \frac{ x-x'}{\ell} \right)^2 \right),
	\end{equation*}
	\noindent where the dependency on the hyperparameters had been omitted. Applying the Bernoulli beam theory to the original GP model, the kernels relating deflections and rotations, strains, bending moments, shear forces and applied loads are obtained, respectively, by:
    \begin{align*}
        &k_{w_b \varphi_b} = \frac{\partial}{\partial x'} k_{w_b w_b}, \\
        &k_{w_b \epsilon_b} = -z \frac{\partial^2}{\partial x'^2} k_{w_b w_b}, \\
        &k_{w_b M} = EI \frac{\partial^2}{\partial x'^2} k_{w_b w_b}, \\
        &k_{w_b V} = EI \frac{\partial^3}{\partial x'^3} k_{w_b w_b}, \\
        &k_{w_b q} = EI \frac{\partial^4}{\partial x'^4} k_{w_b w_b}. 
	\end{align*}
	The covariance kernels that model the relation between rotations in the Euler-Bernoulli model with the other physical quantities are similarly obtained by:
	\begin{align*}
        &k_{\varphi_b w_b} = \frac{\partial}{\partial x} k_{w_b w_b}, \\
        &k_{\varphi_b \varphi_b} = \frac{\partial}{\partial x} \frac{\partial}{\partial x'} k_{w_b w_b}, \\
        &k_{\varphi_b \epsilon_b} = - z\frac{\partial}{\partial x} \frac{\partial^2}{\partial x'^2} k_{w_b w_b}, \\
        &k_{\varphi_b M} = EI \frac{\partial}{\partial x} \frac{\partial^2}{\partial x'^2} k_{w_b w_b}, \\
        &k_{\varphi_b V} = EI \frac{\partial}{\partial x} \frac{\partial^3}{\partial x'^3} k_{w_b w_b},\\
        &k_{\varphi_b q} = EI \frac{\partial}{\partial x} \frac{\partial^4}{\partial x'^4} k_{w_b w_b}. 
	\end{align*}
	The strains are linear along the height of the cross-section, and their kernels are obtained by:
	\begin{align*}
        &k_{\epsilon_b w_b} = -z \frac{\partial^2}{\partial x^2} k_{w_b w_b}, \\
        &k_{\epsilon_b \varphi_b} = -z \frac{\partial^2}{\partial x^2} \frac{\partial}{\partial x'} k_{w_b w_b}, \\
        &k_{\epsilon_b \epsilon_b} = z^2 \frac{\partial^2}{\partial x^2} \frac{\partial^2}{\partial x'^2} k_{w_b w_b}, \\
        &k_{\epsilon_b M} = -z EI \frac{\partial^2}{\partial x^2} \frac{\partial^2}{\partial x'^2} k_{w_b w_b} , \\
        &k_{\epsilon_b V} = -z EI \frac{\partial^2}{\partial x^2} \frac{\partial^3}{\partial x'^3} k_{w_b w_b},\\
        &k_{\epsilon_b q} = -z EI \frac{\partial^2}{\partial x^2} \frac{\partial^4}{\partial x'^4} k_{w_b w_b}. 
	\end{align*}
    \noindent The bending moment-related kernels incorporate the bending stiffness $EI$ as:
	\begin{align*}
        &k_{M \! w_b} = EI \frac{\partial^2}{\partial x^2} k_{w_b w_b}, \\
        &k_{M \! \varphi_b} = EI \frac{\partial^2}{\partial x^2} \frac{\partial}{\partial x'} k_{w_b w_b}, \\
        &k_{M \! \epsilon_b} = - z EI \frac{\partial^2}{\partial x^2} \frac{\partial^2}{\partial x'^2} k_{w_b w_b}, \\
        &k_{M \! M} = EI^2 \frac{\partial^2}{\partial x^2} \frac{\partial^2}{\partial x'^2} k_{w_b w_b}, \\
        &k_{M \! V} = EI^2 \frac{\partial^2}{\partial x^2} \frac{\partial^3}{\partial x'^3} k_{w_b w_b},\\
        &k_{M \! q} = EI^2 \frac{\partial^2}{\partial x^2} \frac{\partial^4}{\partial x'^4} k_{w_b w_b}. 
	\end{align*}
	The shear force covariance kernel and its relations to the remaining physical quantities are defined by:
	\begin{align*}
        &k_{V \! w_b} = EI \frac{\partial^3}{\partial x^3} k_{w_b w_b}, \\
        &k_{V \! \varphi_b} = EI \frac{\partial^3}{\partial x^3} \frac{\partial}{\partial x'} k_{w_b w_b}, \\
        &k_{V \! \epsilon_b} = - z EI \frac{\partial^3}{\partial x^3} \frac{\partial^2}{\partial x'^2} k_{w_b w_b}, \\
        &k_{V \! M} = EI^2 \frac{\partial^3}{\partial x^3} \frac{\partial^2}{\partial x'^2} k_{w_b w_b}, \\
        &k_{V \! V} = EI^2 \frac{\partial^3}{\partial x^3} \frac{\partial^3}{\partial x'^3} k_{w_b w_b},\\
        &k_{V \! q} = EI^2 \frac{\partial^3}{\partial x^3} \frac{\partial^4}{\partial x'^4} k_{w_b w_b}, 
	\end{align*}
	\noindent and the applied load covariance models and their links to the internal forces and beam responses are given as:
	\begin{align*}
        &k_{q w_b} = EI \frac{\partial^4}{\partial x^4} k_{w_b w_b}, \\
        &k_{q \varphi_b} = EI \frac{\partial^4}{\partial x^4} \frac{\partial}{\partial x'} k_{w_b w_b}, \\
        &k_{q \epsilon_b} = - z EI \frac{\partial^4}{\partial x^4} \frac{\partial^2}{\partial x'^2} k_{w_b w_b}, \\
        &k_{q M} = EI^2 \frac{\partial^4}{\partial x^4} \frac{\partial^2}{\partial x'^2} k_{w_b w_b}, \\
        &k_{q V} = EI^2 \frac{\partial^4}{\partial x^4} \frac{\partial^3}{\partial x'^3} k_{w_b w_b}, \\
        &k_{q q} = EI^2 \frac{\partial^4}{\partial x^4} \frac{\partial^4}{\partial x'^4} k_{w_b w_b}. 
	\end{align*}
	To conform with the Timoshenko beam theory, the covariance kernels derived need to be modified to incorporate cross-section rotations due to shear effects (c.f. Eq. \ref{eq:TimoFullRot}). The cross-covariance kernels between Timoshenko's rotations and Bernoulli's model are obtained by:
	\begin{align*}
        &k_{\varphi w_b} = \frac{\partial}{\partial x} k_{w_b w_b} - \frac{EI}{kGA} \frac{\partial^3}{\partial x^3} k_{w_b w_b}, \\
        &k_{\varphi \varphi_b} =  \frac{\partial}{\partial x} \frac{\partial}{\partial x'} k_{w_b w_b} - \frac{EI}{kGA} \frac{\partial^3}{\partial x^3} \frac{\partial}{\partial x'} k_{w_b w_b}, \\
        &k_{\varphi \epsilon_b} = -z  \frac{\partial}{\partial x} \frac{\partial^2}{\partial x'^2} k_{w_b w_b} + \frac{zEI}{kGA} \frac{\partial^3}{\partial x^3} \frac{\partial^2}{\partial x'^2} k_{w_b w_b}, \\
        &k_{\varphi \! M} = EI\frac{\partial}{\partial x} \frac{\partial^2}{\partial x'^2} k_{w_b w_b} - \frac{EI^2}{kGA} \frac{\partial^3}{\partial x^3} \frac{\partial^2}{\partial x'^2} k_{w_b w_b}, \\
        &k_{\varphi \! V} = EI \frac{\partial}{\partial x} \frac{\partial^3}{\partial x'^3} k_{w_b w_b} - \frac{EI^2}{kGA} \frac{\partial^3}{\partial x^3} \frac{\partial^3}{\partial x'^3} k_{w_b w_b}, \\
        &k_{\varphi q} = EI \frac{\partial}{\partial x} \frac{\partial^4}{\partial x'^4} k_{w_b w_b} - \frac{EI^2}{kGA} \frac{\partial^3}{\partial x^3} \frac{\partial^4}{\partial x'^4} k_{w_b w_b},
	\end{align*}
	\noindent while the covariance kernels $k_{i \varphi}$, for $i \in \lbrace w_b, \varphi_b, \epsilon_b, M, V, q\rbrace$ are obtained by adjusting the derivative operators. The covariance kernel for Timoshenko's rotation field is finally calculated as follows:
	\begin{align*}
        k_{\varphi \varphi} =& \frac{\partial}{\partial x} \frac{\partial}{\partial x'} k_{w_b w_b} - \frac{EI}{kGA} \frac{\partial}{\partial x} \frac{\partial^3}{\partial x'^3} k_{w_b w_b} \\ 
        &- \frac{EI}{kGA} \frac{\partial^3}{\partial x^3} \frac{\partial}{\partial x'} k_{w_b w_b} + \frac{EI^2}{kGA^2} \frac{\partial^3}{\partial x^3} \frac{\partial^3}{\partial x'^3} k_{w_b w_b}.
	\end{align*}
	Similarly, the deflections in Timoshenko's model can be obtained by integrating the rotation kernels:
	\begin{align*}
        &k_{w w_b} = k_{w_b w_b} - \frac{EI}{kGA} \frac{\partial^2}{\partial x^2} k_{w_b w_b}, \\
        &k_{w \varphi_b} =  \frac{\partial}{\partial x'} k_{w_b w_b} - \frac{EI}{kGA} \frac{\partial^2}{\partial x^2} \frac{\partial}{\partial x'} k_{w_b w_b}, \\
        &k_{w \epsilon_b} = -z  \frac{\partial^2}{\partial x'^2} k_{w_b w_b} + \frac{zEI}{kGA} \frac{\partial^2}{\partial x^2} \frac{\partial^2}{\partial x'^2} k_{w_b w_b}, \\
        &k_{w \! M} = EI \frac{\partial^2}{\partial x'^2} k_{w_b w_b} - \frac{EI^2}{kGA} \frac{\partial^2}{\partial x^2} \frac{\partial^2}{\partial x'^2} k_{w_b w_b}, \\
        &k_{w \! V} = EI \frac{\partial^3}{\partial x'^3} k_{w_b w_b} - \frac{EI^2}{kGA} \frac{\partial^2}{\partial x^2} \frac{\partial^3}{\partial x'^3} k_{w_b w_b}, \\
        &k_{w q} = EI \frac{\partial^4}{\partial x'^4} k_{w_b w_b} - \frac{EI^2}{kGA} \frac{\partial^2}{\partial x^2} \frac{\partial^4}{\partial x'^4} k_{w_b w_b},
	\end{align*}
	\noindent and the covariance kernel of the Timoshenko's deflection field is calculated as:
	\begin{align*}
			k_{w w} = & k_{w_b w_b} - \frac{EI}{kGA} \frac{\partial^2}{\partial x'^2} k_{w_b w_b} - \frac{EI}{kGA} \frac{\partial^2}{\partial x^2} k_{w_b w_b} \\ 
            & + \frac{EI^2}{kGA^2} \frac{\partial^2}{\partial x^2} \frac{\partial^2}{\partial x'^2} k_{w_b w_b}.
	\end{align*}
	The strain kernels for the Timoshenko model are obtained by differentiating the rotation kernels:
    \begin{align*}
        &k_{\epsilon w_b} = -z\frac{\partial^2}{\partial x^2} k_{w_b w_b} + \frac{zEI}{kGA} \frac{\partial^4}{\partial x^4} k_{w_b w_b}, \\
        &k_{\epsilon \varphi_b} =  -z\frac{\partial^2}{\partial x^2} \frac{\partial}{\partial x'} k_{w_b w_b} + \frac{zEI}{kGA} \frac{\partial^4}{\partial x^4} \frac{\partial}{\partial x'} k_{w_b w_b}, \\
        &k_{\epsilon \epsilon_b} = z^2  \frac{\partial^2}{\partial x^2} \frac{\partial^2}{\partial x'^2} k_{w_b w_b} - \frac{z^2EI}{kGA} \frac{\partial^4}{\partial x^4} \frac{\partial^2}{\partial x'^2} k_{w_b w_b}, \\
        &k_{\epsilon \! M} = -zEI\frac{\partial^2}{\partial x^2} \frac{\partial^2}{\partial x'^2} k_{w_b w_b} + \frac{z EI^2}{kGA} \frac{\partial^4}{\partial x^4} \frac{\partial^2}{\partial x'^2} k_{w_b w_b}, \\
        &k_{\epsilon \! V} = -zEI \frac{\partial^2}{\partial x^2} \frac{\partial^3}{\partial x'^3} k_{w_b w_b} + \frac{z EI^2}{kGA} \frac{\partial^4}{\partial x^4} \frac{\partial^3}{\partial x'^3} k_{w_b w_b}, \\
        &k_{\epsilon q} = -zEI\frac{\partial^2}{\partial x^2} \frac{\partial^4}{\partial x'^4} k_{w_b w_b} + \frac{z EI^2}{kGA} \frac{\partial^4}{\partial x^4} \frac{\partial^4}{\partial x'^4} k_{w_b w_b},
	\end{align*}
	\noindent and the covariance kernel of the Timoshenko's strain field is calculated as:
	\begin{align*}
        k_{\epsilon \epsilon} =& z^2 \frac{\partial^2}{\partial x^2} \frac{\partial^2}{\partial x'^2} k_{w_b w_b} - \frac{z^2EI}{kGA} \frac{\partial^2}{\partial x^2} \frac{\partial^4}{\partial x'^4} k_{w_b w_b} \\ 
        &- \frac{z^2EI}{kGA} \frac{\partial^4}{\partial x^4} \frac{\partial^2}{\partial x'^2} k_{w_b w_b} + \frac{z^2 EI^2}{kGA^2} \frac{\partial^4}{\partial x^4} \frac{\partial^4}{\partial x'^4} k_{w_b w_b}.
	\end{align*}
	Lastly, the correlations between Timoshenko's model for deflections, rotations and strains are given by:
	\begin{align*}
        k_{w \varphi} =& \frac{\partial}{\partial x'} k_{w_b w_b} - \frac{EI}{kGA} \frac{\partial^2}{\partial x^2} \frac{\partial}{\partial x'} k_{w_b w_b}  \\ 
        &- \frac{EI}{kGA} \frac{\partial^3}{\partial x'^3} k_{w_b w_b} + \frac{EI^2}{kGA^2} \frac{\partial^2}{\partial x^2} \frac{\partial^3}{\partial x'^3} k_{w_b w_b},\\
        k_{w \epsilon} =& -z  \frac{\partial^2}{\partial x'^2} k_{w_b w_b} + \frac{zEI}{kGA} \frac{\partial^2}{\partial x^2} \frac{\partial^2}{\partial x'^2} k_{w_b w_b} \\ 
        &+ \frac{zEI}{kGA} \frac{\partial^4}{\partial x'^4} k_{w_b w_b} - \frac{z EI^2}{kGA^2} \frac{\partial^2}{\partial x^2} \frac{\partial^4}{\partial x'^4} k_{w_b w_b}, \\
        k_{\varphi \epsilon} =& -z \frac{\partial}{\partial x} \frac{\partial^2}{\partial x'^2} k_{w_b w_b} + \frac{zEI}{kGA} \frac{\partial}{\partial x} \frac{\partial^4}{\partial x'^4} k_{w_b w_b} \\ 
        &+ \frac{zEI}{kGA}\frac{\partial^3}{\partial x^3} \frac{\partial^2}{\partial x'^2} k_{w_b w_b}  - \frac{zEI^2}{kGA^2} \frac{\partial^3}{\partial x^3} \frac{\partial^4}{\partial x'^4} k_{w_b w_b},
	\end{align*}
	\noindent while $k_{\varphi w}$, $k_{\epsilon w}$, $k_{\epsilon \varphi}$ are obtained similarly.
	
\end{document}